%% file: main.tex
\documentclass{article}
\usepackage[
    version=1,
    shorthands,
    savespace,
    ]{preamble}

\usepackage{iclr2026_conference,times}

\title{Interpretable Clinical Classification with Kolmogorov-Arnold Networks}

\author{
    \centerline{Alejandro Almodóvar\orcidlink{0009-0006-0900-4026}\thanks{Corresponding author \texttt{alejandro.almodovar@upm.es}} \quad Patricia A. Apellániz\orcidlink{0000-0002-8604-9758} \quad Alba Garrido\orcidlink{0009-0006-7238-1473}}\\[1.25ex]
    \centerline{\textbf{Fernando Fernández-Salvador}\orcidlink{0009-0006-1000-4146}  \quad \textbf{Santiago Zazo}\orcidlink{0000-0001-9073-7927} \quad \textbf{Juan Parras}\orcidlink{0000-0002-7028-3179}} \\ [4ex]
    \centerline{Information Processing and Telecommunications Center, ETSI de Telecomunicación}\\
    \centerline{Universidad Politécnica de Madrid}\\
    \centerline{Av. Complutense, 30, 28040}\\
    \centerline{Madrid, Spain} \\
}

\begin{document}

\maketitle

\doparttoc 
\faketableofcontents

\begin{abstract}

\textbf{Background and Objective:}
The increasing use of machine learning in clinical decision support has been limited by the lack of transparency of many high-performing models. In clinical settings, predictions must be interpretable, auditable, and actionable. This study investigates \textit{Kolmogorov-Arnold Networks} as intrinsically interpretable alternatives to conventional black-box models for clinical classification on tabular health data, with the objective of combining predictive performance and clinically meaningful transparency.

\textbf{Methods:}
We introduce two \textit{Kolmogorov-Arnold network}-based classification models: the \textit{Logistic Kolmogorov-Arnold Network}, a flexible generalization of logistic regression, and the \textit{Kolmogorov-Arnold Additive Model}, an additive variant that yields transparent symbolic representations through feature-wise decomposability. Both models are evaluated on multiple public clinical datasets and compared with standard linear, tree-based, and neural baselines.

\textbf{Results:}
Across all datasets, the proposed models achieve predictive performance comparable to or exceeding that of commonly used baselines while remaining fully interpretable. \textit{Logistic-Kolmogorov Arnold Network} obtains the highest overall ranking across evaluation metrics, with a mean reciprocal rank of $0.76$, indicating consistently strong performance across tasks. \textit{Kolgorov-Arnold Additive Model} provides competitive accuracy while offering enhanced transparency through feature-wise decomposability, patient-level visualizations, and nearest-patient retrieval, enabling direct inspection of individual predictions.

\textbf{Conclusions:}
\textit{Kolmogorov-Arnold network}-based classifiers provide a practical and trustworthy alternative to black-box models for clinical classification, offering a strong balance between predictive performance and interpretability for clinical decision support. By enabling transparent, patient-level reasoning and clinically actionable insights, the proposed models represent a promising step toward trustworthy artificial intelligence systems suitable for real-world healthcare applications. The code supporting this study is publicly available at \codeurl.
\end{abstract}


\clearpage

\input{1_introduction}

\input{2_methods}

\input{3_results}

\input{4_discussion}
\input{7_statements}
\input{8_acknowledgements}

\bibliography{main}
\bibliographystyle{abbrvnat}
\appendix
\renewcommand{\partname}{}  
\clearpage
\part{Appendix} 

\input{1_hyperparameter_optimization}
\input{2_statistical_tests}
\input{3_more_results}

\end{document}

%% file: 1_introduction.tex
\section{Introduction}
\label{sec:introduction}

Would you take a medication if your doctor could not explain how it works? The same question now applies to Artificial Intelligence (AI). 

As machine learning continues to change clinical workflows (by enhancing diagnostic accuracy, predicting patient outcomes, and enabling treatment personalization), trust in these systems becomes as important as their accuracy, particularly in high-stakes clinical decision-making contexts \citep{bindra2024artificial, xie2025evolution}.

Despite their success, many AI models remain ``black boxes'', that is, systems whose internal decision-making mechanisms are not comprehensible to humans \citep{xu2024medical}. Deep neural networks, in particular, offer state-of-the-art performance but little insight into how predictions are made. This opacity is problematic in high-stakes domains, such as healthcare, where decisions must be explained, justified, and trusted. Clinicians are legally and ethically required to interpret and defend their actions, and the same standard should apply to AI systems \citep{ghassemi2021false, frasca2024explainable}.

To address these challenges, the field of Explainable Artificial Intelligence (XAI) has emerged, with tools like SHapley Additive exPlanations (SHAP) \citep{lundberg2017unified, mosca2022shap}, Local Interpretable Model-Agnostic Explanations (LIME) \citep{ribeiro2016should}, and Shapley Additive Global Importance (SAGE) \citep{covert2020understanding} providing post-hoc explanations. Specific methods, such as Grad-CAM, have been developed to highlight salient regions in medical images \citep{selvaraju2020grad}. These methods are useful, but external to the model: they approximate explanations rather than reflecting the true reasoning of the system \citep{alaa2019demystifying}. In healthcare, this distinction is critical. When lives are on the line, interpretability must be intrinsic, reliable, and human-centered \citep{lipton2018mythos, chen2022explainable, pahud2024orchestrating, alkhanbouli2025role}. Bridging the gap between the objectives of AI developers and the expectations of interpretability from clinical end-users is essential for real-world adoption of machine learning in clinical practice \citep{bienefeld2023solving}. These insights suggest that achieving interpretability is as much a matter of interface and usability as it is of model structure. New approaches are needed to deliver interpretable, technically sound, and clinically meaningful systems.

\begin{figure}[t]
    \centering
    \includegraphics[width=0.9\linewidth]{classification_kans_scheme.drawio.png}
    \caption{Complete scheme of the interpretable classification pipeline (\cref{sec:methods}), including benchmarking experiments (\cref{sec:ml_res}) and interpretablitity results (\cref{sec:xai_res}).}
    \label{fig:scheme}
\end{figure}

A promising direction is to build models that are interpretable by design. That is, models whose internal structure allows direct understanding of their decisions, without relying on heuristics. In this context, Kolmogorov-Arnold Networks (KANs) have recently emerged as such an alternative \citep{liu2024kan, liu2024kan2}. Grounded in function approximation theory, KANs learn mappings via compositions of univariate spline functions. This allows not only powerful approximation capabilities but also the extraction of symbolic formulas that describe how inputs relate to outputs, offering built-in transparency that traditional neural networks lack. Based on classical results in function approximation, KANs learn mappings through compositions of learnable, low-dimensional interpolation functions, which makes them not only powerful approximators, but also inherently interpretable. Unlike standard neural networks, such as Multi-Layer Perceptrons (MLPs), KANs can express the learned function as a symbolic mathematical formula, providing direct insight into the model's decision-making process. This built-in interpretability enables a more transparent understanding of model behavior, eliminating the need for external explanation tools, although these tools can still be applied if desired. Nevertheless, comparisons between KANs and MLPs suggest that while KANs offer notable interpretable advantages, their predictive performance still has room for improvement in some scenarios \citep{yu2024kan}.

KANs have rapidly evolved, incorporating ideas from deep learning such as dropout \citep{altarabichi2024dropkan}, residual connections \citep{yu2024residual}, transformer-based architectures \citep{genet2024temporal}, convolutional operators \citep{bodner2024convolutional, drokin2024kolmogorov}, U-Net structures \citep{li2024u}, graph neural networks \citep{zhang2024graphkan}, Gaussian processes \citep{chen2024gaussian}, and reinforcement learning frameworks \citep{kich2024kolmogorov}. However, their application to clinical decision-support remains largely unexplored. Most prior work has focused on regression tasks in physics and engineering \citep{ji2024comprehensive, somvanshi2024survey}, or on structured prediction in non-clinical domains. Even in recent medical applications, such as medical imaging \citep{yang2025medkan}, the potential of KANs for interpretable classification in tabular health data is largely unexplored. In particular, it remains unclear how the structure of KANs can be leveraged to meet the transparency requirements of real-world clinical workflows. This gap is especially relevant given that many medical AI applications involve binary or multiclass classification problems (e.g., diagnosis, triage, or risk prediction) where interpretability directly influences clinical trust, adoption, and applicability.

In this study, we address this gap by building upon the KAN framework to propose two novel models for classification on tabular medical data: Logistic Kolmogorov-Arnold Network (\emph{Logistic-KAN}) and Kolmogorov-Arnold Additive Model (\emph{KAAM}). Inspired by Neural Additive Models (NAMs) \citep{agarwal2021neural}, our approach is designed to combine high predictive performance with intrinsic interpretability, tailored to the needs of clinical users (see \cref{fig:scheme}). \emph{Logistic-KAN} integrates the flexibility of KANs into a well-known model structure, enabling nonlinear yet interpretable transformations of each input feature. KAAM further extends this framework by enforcing an additive decomposition, where each variable contributes independently through a dedicated KAN block. This design supports intuitive and localized interpretations while allowing clinicians to explore how individual features influence predictions. Beyond predictive performance, \emph{KAAM} also enables a set of graphical visualization tools (including radar plots, feature contribution visualizations, and nearest-patient comparisons) that facilitate the understanding of individual predictions, particularly in ambiguous or counterintuitive cases. Through extensive experimentation, we show that both models achieve competitive results while maintaining high levels of transparency. These contributions position \emph{Logistic-KAN} and \emph{KAAM} as promising tools for advancing interpretable, trustworthy AI systems in healthcare, combining symbolic transparency with a practical interpretability framework enriched by patient-level visualizations and clinically grounded reasoning.

%% file: 2_methods.tex
\section{Methods}
\label{sec:methods}

\subsection{Mathematical Background}
We consider a classification problem in a clinical setting involving tabular data. In this setup, we assume a cohort of $N$ patients, each described by a medical record consisting of $M$ covariates, which may be of different types (e.g., binary, categorical, or continuous). These covariates represent clinically relevant variables, such as demographic information, laboratory results, or omic features. 

Let $x_i^j$ denote the value of covariate $j \in \{1, 2, \dots, M\}$ for patient $i \in \{1, 2, \dots, N\}$. Then, the $M$-dimensional vector of covariates of patient $i$ is $x_i$, and the dataset is represented as an $N \times M$ matrix $X$, where each row corresponds to a patient. If we refer to a specific covariate $j$ across all patients, we denote it as $x^j$. The associated label for patient $i$ is denoted by $y_i \in \{0, 1, \dots, P-1\}$, where $P$ is the number of possible classes. In many cases, the task is binary classification ($P=2$), but our proposed approach also accommodates multiclass settings ($P > 2$).

\subsubsection{Neural Networks}
A neural network is composed of stacked layers of neurons, where each neuron performs a nonlinear transformation of its input. Specifically, a neuron $k$ computes:
\begin{equation}
    \label{eq:nn_layer}
    z_k = \chi \left( \omega_k^\top v \right),
\end{equation}

where $v$ denotes the input vector to the neuron, $\omega_k$ is the corresponding weight vector (which may include a bias term), $\chi(\cdot)$ is a nonlinear activation function chosen by the designer (e.g., ReLU, sigmoid), and $z_k$ is the scalar output of neuron $k$.

By grouping several such neurons, we form a layer, which maps the input vector $v$ to an output vector $z = [z_1, z_2, \dots, z_K]^\top$. The dimensionality of the input and output vectors can differ depending on the number of neurons in the layer. A deep neural network is then constructed by stacking multiple such layers, where the output of one layer serves as the input to the next. A common architecture is the MLP, which consists of an input layer, one or more hidden layers, and an output layer.

The success of MLPs comes from their ability to approximate complex multivariate functions. Under mild assumptions, MLPs are universal function approximators, which means they can approximate any continuous function \citep{hornik1989multilayer}. The parameters $\omega_k$ of each neuron are typically learned through gradient-based optimization algorithms such as stochastic gradient descent \citep{haykin1994neural}. Once trained, the resulting network defines a nonlinear mapping from inputs to outputs that approximates the underlying structure of the training data.

\subsubsection{Kolmogorov-Arnold Networks}
The mathematical foundation of KANs lies in the Kolmogorov Representation Theorem (KRT) \citep{kolmogorov1961representation, liu2024kan}, which states that any smooth multivariate function $f: [0,1]^M \rightarrow \mathbb{R}$ can be expressed as:

\begin{equation}
    \label{eq:repre_theorem}
    f(x) = f(x^1, ..., x^M) = \sum_{q=1}^{2M+1} \Phi_q \left( \sum_{k=1}^M \phi_{q, k} (x^k) \right),
\end{equation}

where $\phi_{q,k}: [0,1] \rightarrow \mathbb{R}$ and $\Phi_q: \mathbb{R} \rightarrow \mathbb{R}$ are univariate functions. In essence, any $M$-dimensional smooth function can be decomposed into a combination of $M(2M+1)$ univariate functions and summations. Although theoretically elegant, this result was historically regarded as of limited practical utility \citep{girosi1989representation, poggio2020theoretical, schmidt2021kolmogorov}, with few exceptions \citep{kuurkova1991kolmogorov}, especially given the empirical success of neural networks supported by the Universal Approximation Theorem \citep{hornik1989multilayer}.

Recently, Liu et al. \citep{liu2024kan} introduced the KAN, an architecture inspired by \Eqref{eq:repre_theorem}, which has demonstrated competitive performance relative to MLPs. A KAN is built by stacking KAN layers, in analogy with the way deep neural networks are constructed by stacking neuron layers. A KAN layer consists of a matrix $\Phi$ of univariate functions $\phi_{p,k}$, where $p$ indexes the input dimension and $k$ the output dimension. Each function $\phi_{p,k}$ acts as a learnable nonlinear transformation, serving a role similar to activation functions in neural networks. Thus, these models can be viewed as a structural counterpart to MLPs: whereas traditional networks learn the weights of affine transformations followed by fixed nonlinearities, KANs learn the nonlinearities themselves (the ``activations'' in MLPs). The overall function implemented by the network is obtained by composing multiple KAN layers, analogous to deep architectures in standard neural networks.

This parallel extends to the theoretical level. The original KRT in \Eqref{eq:repre_theorem} corresponds to a two-layer KAN, where the first layer has $M$ inputs and $2M+1$ outputs, and the second layer aggregates them into a scalar output. However, as with MLPs, empirical performance benefits from stacking multiple layers. Deep KANs, composed of several such layers, can approximate more complex functions and have shown promising results across different domains \citep{liu2024kan}.

To enable training with modern optimization algorithms, the functions $\phi_{p,k}$ are parameterized using piecewise polynomial basis functions, specifically B-splines. These are equipped with mechanisms such as residual connections and adaptive knot placement \citep{liu2024kan}. Training a KAN involves learning the parameters of these splines to minimize a suitable loss function, and, importantly, each learned spline can be approximated as a symbolic function, yielding an interpretable analytical representation of the learned mapping.

Despite their theoretical appeal and promising empirical results, KANs face several practical challenges. In particular, they tend to exhibit worse scaling behavior than MLPs. For example, while an MLP with depth $L$ and width $N$ requires $O(N^2 L)$ parameters, a KAN with the same structure also depends on the spline polynomial degree $k$ and the number of grid points $G$, leading to a complexity of $O(N^2 L (G + k))$. Although initial studies suggest that KANs may require lower width to achieve comparable performance \citep{liu2024kan}, this remains an open question \citep{yu2024kan}, and the trade-offs between expressivity, interpretability, and scalability are still being actively investigated.

\subsubsection{Additively Separable Functions}
An additively separable function (ASF) \citep{segal1994sufficient} is a multivariate function that can be decomposed as:
\begin{equation}
    \label{eq:gam}
    f(x) = f(x^1, ..., x^M) = \alpha + \sum_{j=1}^M g_j(x^j),
\end{equation}

where each $g_j$ is a univariate (potentially nonlinear) function, commonly referred to as a shape function \citep{lou2012intelligible}, and $\alpha$ is a scalar intercept. Models of this form belong to the family of Generalized Additive Models (GAMs) \citep{hastie1990generalized}, a class of models that combines flexibility and interpretability. Since each $g_j$ may capture nonlinear effects, GAMs are more expressive than linear models, while the additive structure allows for intuitive interpretation of each feature's marginal contribution \citep{marcinkevivcs2023interpretable, lou2012intelligible, lipton2018mythos}.

Due to their balance between interpretability and predictive power, GAMs have been recommended for medical modeling tasks \citep{hastie1995generalized, utkin2022survnam}. Extensions of GAMs have been proposed to improve their performance and usability, including intelligible models based on boosted trees \citep{caruana2015intelligible}, and neural network-based variants such as NAMs \citep{agarwal2021neural}. In parallel, significant effort has been devoted to improving the interpretability of GAMs via visual analytics and model visualization techniques \citep{fasiolo2020scalable, hohman2019telegam, wang2021gam}.

Importantly, KANs can be seen as a generalization of GAMs. Specifically, the formulation in \Eqref{eq:gam} corresponds to a KAN with a single layer, or equivalently, a two-layer KAN where the second layer consists of identity functions. This perspective enables the training of GAM-like models using gradient-based optimization, which can offer advantages over traditional GAM training procedures that rely on iterative smoothing or tree ensembles \citep{agarwal2021neural}.

Moreover, KANs provide two distinct pathways to recover ASFs: either by explicitly designing a shallow KAN with a single layer (which guarantees additivity but lacks universal approximation power as it does not satisfy the full KRT \Eqref{eq:repre_theorem}), or by promoting sparsity in deeper KAN architectures, potentially inducing additivity as an emergent property. While the theoretical expressive power of shallow KANs is limited, empirical evidence suggests that they may suffice for many practical problems. The strong success of NAMs supports this hypothesis \citep{agarwal2021neural}, which imposes additive structure by modeling each $g_j$ with an independent neural network.

In summary, while classical GAMs are constrained both in expressivity and in their reliance on non-differentiable training algorithms, KANs offer a principled and flexible alternative. They retain the interpretability of GAMs, may surpass NAMs in transparency, and allow training via end-to-end differentiable optimization. As GAMs are effectively a subset of KANs, existing interpretation techniques developed for GAMs can be directly adapted and extended to the KAN framework.

\subsubsection{Logistic Regression}
We now consider the classical case of binary classification ($P = 2$), where each label $y_i \in \{0,1\}$ indicates the presence or absence of a clinical condition. As introduced earlier, $x_i$ denotes the $M$-dimensional covariate vector of patient $i$, $X$ is the full $N \times M$ dataset for $N$ patients, and $y \in \{0,1\}^N$ is the corresponding vector of binary outcomes.

Logistic Regression (LR) \citep{hosmer2013applied} is one of the most widely used models in clinical research due to its simplicity. It models the conditional probability of class 1 as:

\begin{equation}
    \label{eq:logistic_regression}
    P(y_i=1 | x_i) = \frac{1}{1 + e^{-\beta^T x_i}} = \sigma(\beta^T x_i),
\end{equation}

where $\beta$ is the $M$-dimensional parameter vector and $\sigma(\cdot)$ is the sigmoid function.

The model is typically trained by minimizing the negative log-likelihood of the Bernoulli distribution:

\begin{equation}
    \label{eq:lr_opt_pose}
    \begin{split}
                \min_{\beta}  \mathcal{L(\beta)} =& \sum_{i=1}^N - \bigg( y_i \log \big( \sigma(\beta^T x_i) \big) + (1-y_i) \log \big( 1 - \sigma(\beta^T x_i) \big) \bigg).
    \end{split}
\end{equation}

The gradient of the objective function \Eqref{eq:lr_opt_pose} can be computed analytically using the derivative of the sigmoid function, $\sigma'(x) = \sigma(x)(1 - \sigma(x))$, leading to:

{\footnotesize
\begin{equation}
    \label{eq:lr_grad}
    \begin{split}
        \nabla_{\beta} \mathcal{L}(\beta) &= \sum_{i=1}^N -y_i \frac{\sigma'(\beta^T x_i)}{\sigma(\beta^T x_i)} x_i + (1-y_i) \frac{\sigma'(\beta^T x_i)}{1 - \sigma(\beta^T x_i)}x_i \\
        \quad&= \sum_{i=1}^N -y_i \big(1 - \sigma(\beta^T x_i) \big)x_i + (1-y_i) \sigma(\beta^T x_i)x_i\\ 
        \quad&= \sum_{i=1}^N \big( -y_i + y_i \sigma(\beta^T x_i) + \sigma(\beta^T x_i) - y_i \sigma(\beta^T x_i) \big) x_i \\ 
        \quad&= \sum_{i=1}^N \big(\sigma(\beta^T x_i) - y_i \big) x_i.
    \end{split}
\end{equation}}

Since this gradient does not admit a closed-form solution, the model is typically optimized using iterative methods such as gradient descent, quasi-Newton, or Newton-based algorithms, which can exploit the analytical expressions of both the gradient and the Hessian.

LR implicitly assumes a linear and additive structure in the logit space. Specifically, we can rewrite \Eqref{eq:logistic_regression} as:

\begin{equation}
    \label{eq:logistic_regression_sum}
    P(y_i=1 | x_i) = \sigma \left( \sum_{j=1}^M \beta_j x_{i}^j \right),
\end{equation}

which highlights that each covariate contributes independently and linearly to the logit. In this sense, LR can be interpreted as a special case of a GAM applied to the logit space, where each shape function $g_j(x^j)$ is constrained to be linear. While this assumption simplifies understanding and training, it also limits expressiveness. In particular, the model cannot capture nonlinear relationships or interactions between features, which are often present in real-world clinical data.

Despite its simplicity, LR remains a widely used model in clinical research and practice \citep{hosmer2013applied}, due to three key advantages: computational efficiency, compatibility with discrete features, and inherent interpretability. First, LR can be trained efficiently via standard optimization methods, as discussed in the previous section. Second, it is particularly well-suited to handling discrete or categorical covariates (common in clinical datasets) through techniques such as one-hot encoding. Although care must be taken to avoid multicollinearity, this encoding strategy effectively linearizes the feature space by enumerating a limited number of categorical configurations, which can be well captured by the model's degrees of freedom. Third, and perhaps most importantly in a medical context, LR offers built-in interpretability. Each coefficient $\beta_j$ directly reflects the contribution of feature $j$ to the log-odds of the predicted probability. When the input matrix $X$ is properly normalized, the magnitude and sign of each component of $\beta$ can be interpreted as the relative importance and direction of influence of the corresponding covariate. This makes LR highly attractive for clinical decision support, as its outputs can be easily communicated to healthcare professionals and stakeholders. Consequently, LR models are ubiquitous in medical applications \citep{hosmer2013applied}. Notable examples include their use in Cox proportional hazards models \citep{cox1972regression} for survival analysis, propensity score modeling \citep{rosenbaum83propensity, garrido2014methods, austin2011introduction}, and the estimation of average treatment effects \citep{austin2010performance, hernan2020causal}.

Nevertheless, the limitations of LR are well known. The model assumes that the log-odds of the outcome are a linear function of the input features, which may not hold in real-world data. In such cases, this structural misspecification can lead to suboptimal predictive performance. A common extension replaces the linear term with a nonlinear function $f_{\omega}$, typically parameterized by a neural network:

\begin{equation}
    \label{eq:mlp_logistic}
    P(y_i=1 | x_i) = \sigma \left( f_{\omega} (x_i) \right),
\end{equation}

where $f_{\omega}$ denotes the neural network with trainable weights $\omega$. This approach retains the probabilistic structure of LR but increases its expressivity through the Universal Approximation Theorem. It is often trained using the same binary cross-entropy loss as LR, corresponding to the negative log-likelihood under a Bernoulli model. However, this flexibility comes at a cost: the resulting model no longer provides interpretable coefficients, and its decision process becomes opaque to end users.

\subsection{Our proposals: \emph{Logistic-KAN} and \emph{KAAM}}
\begin{figure}[!ht]
\centering
\begin{subfigure}{0.49\linewidth}
    \centering
    \includegraphics[width=\linewidth]{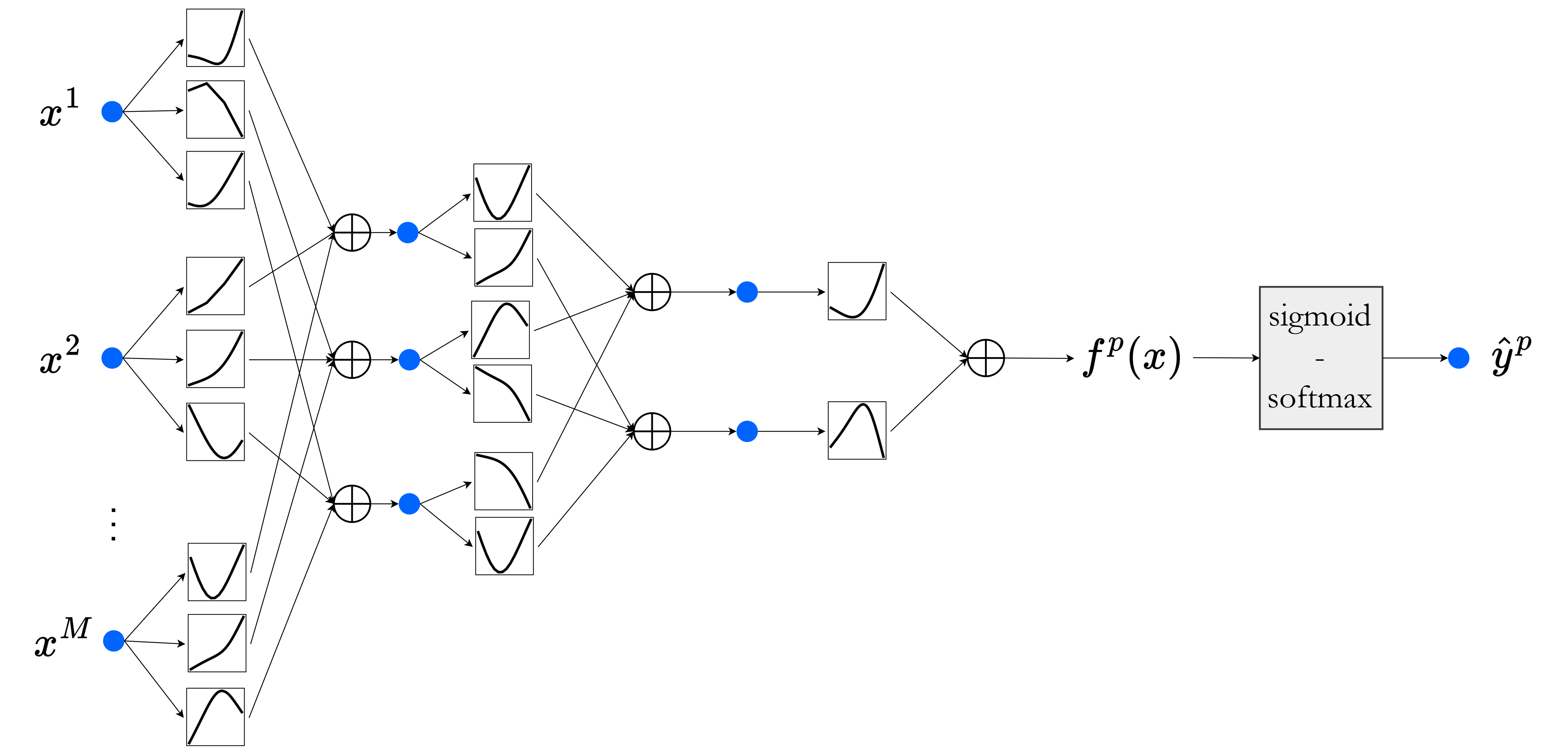}
    \caption{\emph{Logistic-KAN} architecture}
\end{subfigure}
\begin{subfigure}{0.49\linewidth}
    \centering
    \includegraphics[width=\linewidth]{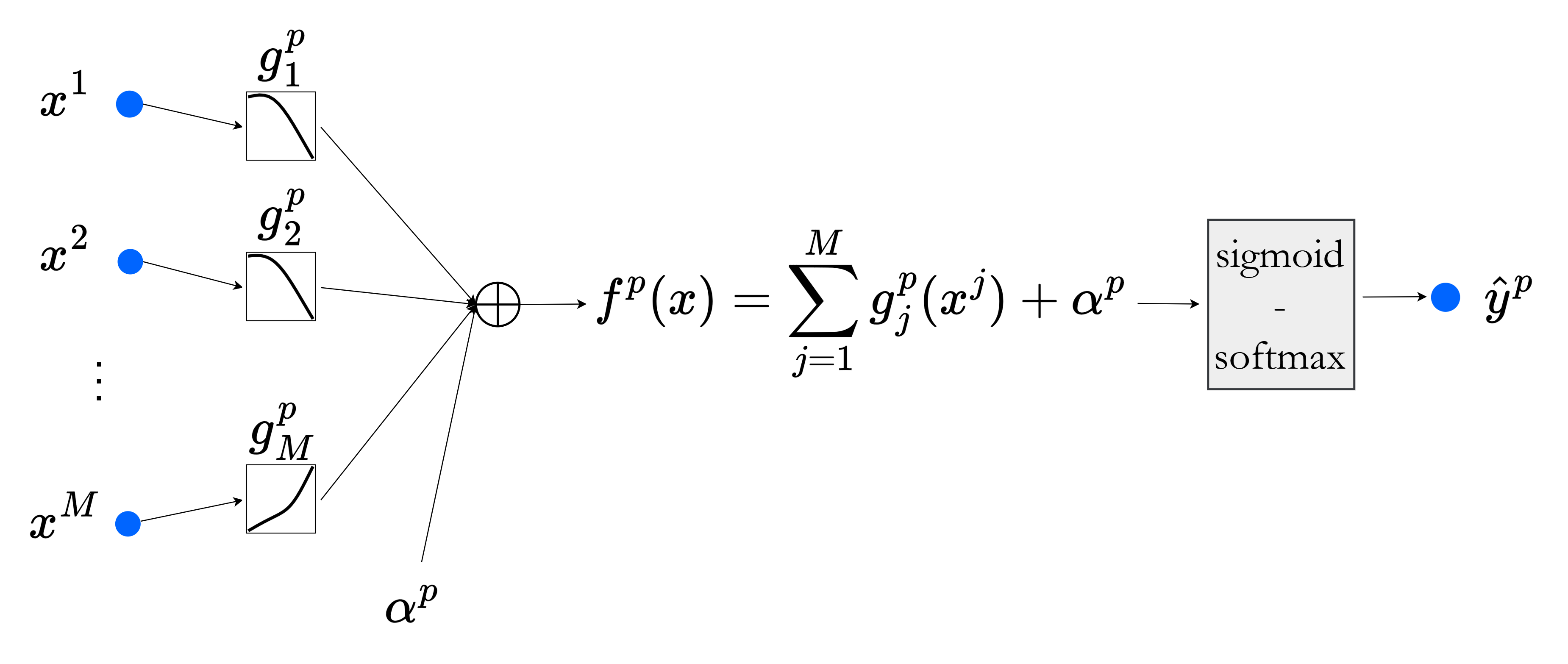}
    \caption{\emph{KAAM} architecture}
\end{subfigure}
\caption{Comparison of the proposed \emph{Logistic-KAN} and \emph{KAAM} models. In \emph{Logistic-KAN}, the full input vector $(x^1, \dots, x^M)$ is processed jointly through a KAN with multiple layers. The resulting function $f_\varphi^p(x)$ allows for flexible, nonlinear interactions between covariates while retaining interpretability through spline-based components. The output logit is then passed through a sigmoid or softmax activation to produce the prediction $\hat{y}^p$.In contrast to \emph{Logistic-KAN}, in \emph{KAAM}, each input covariate $x^j$ is independently transformed by a univariate, class-specific function $g_j^p(x^j)$. These transformed individual contributions are summed additively, along with a bias term $\alpha^p$, to compute the class-specific logit $f^p(x)$. Applying a softmax or sigmoid yields the final prediction $\hat{y}^p$. This additive structure enforces feature-wise decomposability and enhances interpretability.}
\label{fig:architectures}
\end{figure}

Motivated by these trade-offs, we propose two KAN-based models: Logistic Kolmogorov-Arnold Network (\emph{Logistic-KAN}) and Kolmogorov-Arnold Additive Model (\emph{KAAM}). These models aim to preserve the clinical interpretability of LR while expanding its expressive power. The following subsections provide a detailed description of each model. Figure \ref{fig:architectures} provides a high-level visual comparison of both models.

\Cref{tab:model_properties} summarizes the main differences between classical and recent models in terms of structural properties and interpretability. Our proposed models aim to combine the best of both worlds: expressive power and symbolic interpretability.

\begin{table}[!ht]
\centering
\renewcommand{\arraystretch}{1.4}
\resizebox{0.6\textwidth}{!}{
\begin{tabular}{l|cccc}
\hline\hline
\textbf{Model} & \textbf{Linear} & \textbf{Additive} & \textbf{Universal Approximator} & \textbf{Interpretability} \\
\hline\hline
\textit{LR} & yes & yes & no & Advanced \\
\textit{GAM} & no & yes & no & Interpretable \\
\textit{NAM} & no & yes & no & Interpretable \\
\textit{MLP} & no & no & yes & Black-box \\
\textit{Logistic-KAN} & no & no & yes & Interpretable \\
\textit{KAAM} & no & yes & no & Advanced \\
\hline\hline
\end{tabular}
}
\caption{Comparison of the models considered in this study based on four key properties: linearity, additive structure, universal approximation, and interpretability. Models labeled as \textit{Interpretable} provide structural mechanisms for interpretation, such as per-feature shape functions or symbolic components. Models marked as \textit{Advanced} offer enhanced interpretability through explicit decompositions (e.g., logit matrices) and individualized analysis tools, supporting clinically actionable interpretations beyond standard post-hoc techniques.}
\label{tab:model_properties}
\end{table}

\subsubsection{Logistic-KAN: Generalizing logistic regression with interpretability}
To bridge the gap between model flexibility and built-in interpretability, we propose using KANs as the function approximator within a classification framework. Analogous to the neural network model in \Eqref{eq:mlp_logistic}, we define a KAN function $f_{\varphi}(x_i)$ that maps patient covariates $x_i$ to $P$ output logits, where $P$ denotes the number of classes. To obtain class probabilities, we apply the softmax transformation to the output logits:

\begin{equation}
    \label{eq:kan_logistic_multi}
    P(y_i = p \mid x_i) = \frac{e^{f^p_{\varphi}(x_i)}}{\sum_{k=0}^{P-1} e^{f^k_{\varphi}(x_i)}},
\end{equation}

where $f^p_{\varphi}(x_i)$ denotes the $p$-th component of the KAN output. This formulation \Eqref{eq:kan_logistic_multi} guarantees a valid probability distribution over classes, making the model suitable for standard multi-class classification tasks. Training can be performed using conventional techniques, such as minimizing cross-entropy loss, optionally incorporating class weighting to handle imbalanced datasets.

In the binary classification case ($P=2$), the model reduces to a logistic form:

\begin{equation}
    \label{eq:kan_logistic_binary}
    \begin{split}
            P(y_i=1 | x_i) =& \frac{e^{f^1_{\varphi}(x_i)}}{e^{f^0_{\varphi}(x_i)} + e^{f^1_{\varphi}(x_i)}} = \frac{1}{1 + e^{f^0_{\varphi}(x_i) - f^1_{\varphi}(x_i)}} = \sigma \left( f^1_{\varphi}(x_i) - f^0_{\varphi}(x_i) \right),
    \end{split}
\end{equation}

where $\sigma(\cdot)$ is the sigmoid function. This expression closely resembles both the standard LR formulation \Eqref{eq:logistic_regression} and its neural network generalization \Eqref{eq:mlp_logistic}, but replaces the linear or MLP-based function $f$ with a KAN.

From a training perspective, the \emph{Logistic-KAN} model is compatible with the same optimization strategies used for LR and neural networks. In particular, it can be trained via gradient descent to minimize the negative log-likelihood of the Bernoulli or categorical distributions.

Crucially, however, the KAN introduces a significant advantage in terms of expressiveness. While MLPs require post-hoc explanation tools (e.g., SHAP, LIME, or SAGE) to gain insight into predictions, KANs learn explicit spline-based representations that can be directly interpreted. These univariate spline functions can be approximated by symbolic expressions, enabling the extraction of analytical descriptions of the learned decision boundaries. This built-in interpretability allows KANs to retain the flexibility of deep models while simultaneously addressing the transparency demands of clinical applications.

Compared to standard LR, the \emph{Logistic-KAN} offers greater expressiveness by accommodating nonlinear transformations and interactions among covariates. In cases where the data exhibits a strong linear structure, the KAN may converge to a solution similar to that of LR. However, it provides a valuable additional benefit: suppose the learned splines are approximately linear, this outcome serves as evidence that a linear model suffices. In contrast, applying LR a priori lacks this built-in model validation mechanism and requires external comparison with more flexible models to assess fit quality.

\subsubsection{KAAM: Enhancing clinical interpretability through additive modeling}\label{sec:kaam_xai}
While the \emph{Logistic-KAN} model provides an interpretable structure through learnable univariate spline functions, its $f_{\varphi}(x_i)$ form can still be complex, especially in multilayer architectures where interactions between covariates make human interpretation difficult. To address this limitation, we introduce a simplified variant, the \emph{KAAM}, which explicitly enforces an additive structure over input features. This constraint significantly enhances the model’s transparency, making it particularly suitable for clinical applications.

The \emph{KAAM} can be seen as a constrained variant of the \emph{Logistic-KAN}, where the function $f_{\varphi}$ is composed of a single KAN layer and is explicitly defined as an ASF:

\begin{equation}
\label{eq:asf_kan}
f_{\varphi}^p(x_i) = f_{\varphi}^p (x_i^1, x_i^2, ..., x_i^M) = \alpha^p + \sum_{j=1}^M g_j^p(x_i^j),
\end{equation}

where each $g_j^p$ is a univariate spline-based function specific to covariate $j$ and class $p$, and $\alpha^p$ is a class-specific bias term. This structure mirrors the form of a GAM \citep{hastie1990generalized} in \Eqref{eq:gam}, but differs in the function approximator: while GAMs typically rely on manually specified smoothers or decision trees, and NAMs \citep{agarwal2021neural} use small neural networks, our approach leverages KANs to learn each $g_j^p$. This provides the dual benefit of flexible approximation and symbolic interpretability via spline inspection.

The simplicity of \emph{KAAM} comes at a cost in expressivity, as single-layer models do not fulfill the KRT \Eqref{eq:repre_theorem} and hence are not universal approximators. However, we hypothesize that in many clinical applications, such additive models may be sufficient to capture relevant information while offering significantly improved interpretability, particularly when decision transparency is a primary requirement. Our experimental results support this hypothesis.

To formalize this model, we define the $N \times (M+1)$ logit matrix $\Delta_p$ for each class $p$:

\begin{equation}
\label{eq:logit_matrix_multi}
    \Delta_p = \left[\begin{array}{ccccc}
         g_1^p(x^1_1) & g_2^p(x^2_1)  & \cdots & g_M^p(x^M_1) & \alpha^p  \\
         g_1^p(x^2_1) & g_2^p(x^2_2)  & \cdots & g_M^p(x^M_2) & \alpha^p  \\
         \vdots & \vdots  & \ddots & \vdots & \vdots  \\
         g_1^p(x^1_N) & g_2^p(x^2_N)  & ... & g_M^p(x^M_N) & \alpha^p  \\ 
    \end{array} \right].
\end{equation}

Each row of $\Delta_p$ represents the per-feature logit contributions for a patient. Summing across each row gives the total logit $f_{\varphi}^p(x_i)$ for patient $i$, which is then transformed into a probability via the softmax function in \Eqref{eq:kan_logistic_multi}.

In binary classification ($P = 2$), we can simplify the formulation by constructing a single differential logit matrix $\Delta$, where each entry column reflects the contribution of a covariate to the decision margin:

\begin{equation}
    \label{eq:logit_matrix_bin}
    \resizebox{0.9\columnwidth}{!}{$
    \Delta = \left[\begin{array}{ccccc}
         g_1^1(x^1_1) - g_1^0(x^1_1) & g_2^1(x^2_1) - g_2^0(x^2_1)  & \cdots & g_M^1(x^M_1) - g_M^0(x^M_1) & \alpha^1 - \alpha^0  \\
         g_1^1(x^2_1) - g_1^0(x^2_1) & g_2^1(x^2_2) - g_2^0(x^2_2)  & \cdots & g_M^1(x^M_2) - g_M^0(x^M_2) & \alpha^1 - \alpha^0  \\
         \vdots & \vdots  & \ddots & \vdots & \vdots  \\
         g_1^1(x^1_N) - g_1^0(x^1_N) & g_2^1(x^2_N) - g_2^0(x^2_N)  & ... & g_M^1(x^M_N) - g_M^0(x^M_N) & \alpha^1 - \alpha^0  \\ 
    \end{array} \right].
    $}
\end{equation}

Now, the probability of patient $i$ being in class 1 is then obtained by summing row $i$ and applying the sigmoid function.

For further analysis, we define the average contribution $M+1$ vector $\delta$ as the mean of $\Delta$ across patients. This vector summarizes the average effect of each feature (and the bias) across the cohort.

In the following subsections, we exploit the \emph{KAAM} structure and logit decomposition to introduce several visualization tools that facilitate model understanding. These include PRPs of per-feature contributions, feature importance analysis, and nearest-patient comparisons, each of which supports different aspects of clinical interpretability. While our presentation focuses on the binary case for clarity, all tools generalize naturally to multiclass settings by repeating the procedure for each class logit $f^p_{\varphi}(x_i)$.

\paragraph{Partial Dependence Plots.}
One of the most intuitive and widely used tools for model interpretability in GAMs is the Partial Dependence Plot (PDP) \citep{hastie2009elements}. In the context of \emph{KAAMs}, these plots leverage the fact that each function $g_j$ depends only on a single covariate, allowing us to visualize its isolated contribution to the overall prediction. For each input variable $x^j$, we can directly plot the shape function $g_j^p(x^j)$ to observe how changes in the value of that covariate influence the logit for class $p$. This visualization provides a clear understanding of the marginal effect of $x^j$, independent of other features, which is particularly valuable in clinical settings where domain experts often seek to assess individual risk factors.

In binary classification, the difference $g_j^1(x^j) - g_j^0(x^j)$ contributes additively to the logit (i.e., the input to the sigmoid function in \Eqref{eq:kan_logistic_binary}), rather than directly to the predicted probability. Nevertheless, since the sigmoid is a monotonic function, higher logits correspond to higher predicted probabilities. Thus, the qualitative interpretation of PDPs remains valid: increasing $g_j^1(x^j) - g_j^0(x^j)$ generally increases the model’s confidence in class 1.

\paragraph{Feature Importance Analysis.}
The \emph{KAAM} framework also enables a straightforward and interpretable computation of feature importance by leveraging the structure of the logit matrix $\Delta$. Recall that each column of $\Delta$ corresponds to the additive contribution of a single covariate (plus a bias term) across all patients. This structure enables us to assess the variation of each covariate across the population and, consequently, its contribution to discriminating between different prediction outcomes.

In particular, if a column $j$ in $\Delta$ exhibits low variance, it suggests that covariate $j$ contributes similarly across patients and is thus less informative for classification. Conversely, a column with high variance indicates that the covariate plays a significant role in differentiating between low- and high-risk patients. The final bias column (e.g., $\alpha^1 - \alpha^0$ in \Eqref{eq:logit_matrix_bin}) is constant by construction and serves as a baseline reference. We quantify feature importance by computing the empirical variance of each column in $\Delta$. This results in an intuitive, model-driven ranking of covariates based on their contribution to the prediction logit.

Crucially, this importance measure is global, as it is computed across the entire cohort. However, the same analysis can be applied to any subpopulation of interest (e.g., high-risk patients, specific age groups, treatment cohorts), enabling a more granular understanding of how the model's reliance on specific features varies across patient subgroups. This facilitates subgroup-aware auditing of the model, supporting trust and transparency in sensitive clinical contexts.

\paragraph{Probability Radar Plots.}
A Probability Radar Plot (PRP) is a radial visualization tool that facilitates individualized model interpretation \citep{saary2008radar}. In the context of \emph{KAAMs}, we adapt this tool to highlight the impact of each covariate on the predicted probability for a specific patient $i$, by comparing it against an ``average patient'' baseline $\sigma(\delta)$. We define the average patient as the one whose input covariates correspond to the mean logit contributions across the population, i.e., the vector $\delta = [\delta^1, \delta^2, \dots, \delta^M, \alpha]$ computed from the mean of the logit matrix $\Delta$. The associated predicted probability is $\sigma(\alpha + \sum_{j=1}^M \delta^j)$.

To assess how an individual patient $i$ deviates from this baseline, we construct a PRP where each ray corresponds to a single covariate $x^j$. For each $j$, we compute the probability obtained when replacing the $j$-th logit contribution in the average vector with that of the actual patient:

\begin{equation}
\sigma\left(\alpha + \sum_{k \neq j} \delta^k + g_j(x_i^j)\right).
\end{equation}

This value reflects the predicted probability that the average patient would have if only covariate $j$ were replaced by its value in patient $i$. The resulting plot provides a concise and intuitive view of how each covariate influences the prediction, pushing it above or below the baseline. Rays pointing outward indicate covariates that increase the predicted risk compared to the average patient, while rays pointing inward reflect protective or neutral effects.

This analysis can be customized further by redefining the average vector $\delta$ based on clinically meaningful subgroups, such as age strata, comorbidities, or risk profiles. Doing so allows for context-aware comparisons, for instance, evaluating a patient relative to others with the same disease stage or demographic background.

\paragraph{Nearest patient retrieval.}
An additional advantage of the \emph{KAAM} framework is the availability of the logit matrix $\Delta$, which enables patient-level comparisons in a clinically meaningful representation space. In particular, we propose identifying the most similar patients not in the original covariate space (which may fail to reflect the clinical relevance of features), but in the logit space, where covariates are weighted according to their learned predictive importance.

Given a new patient $x_*$, we compute their logit representation $\delta_*$ by evaluating the \emph{KAAM} model on their covariates, i.e., retrieving the row of $\Delta$ from \Eqref{eq:logit_matrix_bin} corresponding to $x_*$. We then measure similarity between $\delta_*$ and the rows of $\Delta$ from the training set using a distance metric. In our experiments, we employ the Euclidean distance, but other metrics could be used depending on the application.

This approach naturally accounts for the model's implicit feature weighting. For instance, in a diabetes classification task, if the \emph{KAAM} has learned that BMI is more influential than sex, then differences in BMI will contribute more heavily to the distance in logit space, effectively prioritizing clinically relevant similarity. From a clinical perspective, retrieving similar patients in logit space offers practical benefits. Physicians can inspect the medical records of their nearest neighbors to inform decision-making, identify treatments that have proven effective in similar cases, or anticipate potential complications. As such, this strategy provides a transparent and actionable tool for case-based reasoning in clinical support systems.

%% file: 3_results.tex
\section{Results}
\label{sec:results}

To assess the full potential of our proposed models in clinical settings, this section is structured into three main parts. First, in Section \ref{sec:data}, we introduce the health-related datasets used in our experiments. Then, in Section \ref{sec:ml_res}, we evaluate the predictive performance of \emph{Logistic-KAN} and \emph{KAAM} on a variety of classification tasks and compare them with state-of-the-art baselines. This step establishes that our models are competitive in terms of machine learning utility. Once predictive performance is established, the focus in this section shifts to the core advantage of our approach: built-in interpretability. We illustrate the native visual interpretability of \emph{KAAM} through a representative multiclass example (Diabetes-130). Finally, in Section \ref{sec:usecase}, we present a comprehensive case study on the Heart dataset, demonstrating how symbolic formulas, personalized reasoning, and patient similarity retrieval can support transparent, clinician-facing decision-making. This ability to deliver accurate and inherently interpretable predictions is especially critical in high-stakes domains, such as healthcare. All datasets and source code used in this study are publicly available at \url{https://github.com/Patricia-A-Apellaniz/classification_with_kans}.

\subsection{Experimental Data: Patient Cohorts and Tasks}\label{sec:data}
To evaluate the effectiveness of our proposed models in diverse clinical scenarios, we conducted experiments on six publicly available health-related datasets. These datasets span a range of prediction tasks, from binary risk classification to multiclass diagnosis estimation, and include real-world records collected from surveys, hospitals, and genomic resources. Importantly, they also reflect realistic conditions commonly found in healthcare data, such as varying levels of class imbalance, from severely skewed distributions (e.g., Heart) to more balanced ones (e.g., Obesity). This diversity makes the evaluation particularly relevant for real-world deployment scenarios. All datasets were preprocessed using a unified pipeline: categorical variables were one-hot encoded, and continuous features were standardized. 

The datasets include: (i) the Heart dataset \citep{cdc_brfss_2022}, whose task is binary classification, predicting presence of heart disease, with only 9.4\% positives; (ii) Diabetes-H, from the CDC BRFSS 2015 \citep{xie2019building}, a multiclass task with three labels: no diabetes, prediabetes, and diabetes; (iii) Diabetes-130 \citep{diabetes_130-us_hospitals_for_years_1999-2008_296}, based on hospital admissions, with demographic and treatment features and a multiclass task of three diagnostic outcomes; (iv) the Obesity dataset \citep{palechor2019dataset}, a 7-class problem ranging from insufficient weight to obesity type III based on lifestyle and physical metrics from Latin American populations; (v) a binarized version of the same, Obesity-Bin, grouping patients as obese or not; and (vi) the Breast Cancer dataset from TCGA \citep{weinstein2013cancer}, with 569 cases and 30 clinical and genomic variables for predicting malignancy. These datasets provide a robust benchmark to evaluate both the predictive and interpretable capabilities of our models across diverse clinical settings. A summary of key dataset characteristics is provided in Table \ref{tab:datasets}.

\begin{table*}[!h]
\centering
\renewcommand{\arraystretch}{1.75}
\resizebox{\textwidth}{!}{
\begin{tabular}{l|cccccc}
\hline\hline
\textbf{Dataset} & \textbf{\# Samples used (available)} & \textbf{\# Features} & \textbf{Data Types} & \textbf{Task} & \textbf{Classes} & \textbf{Class Proportion} \\
\hline\hline
\textit{Heart} & 1,000 (253,680) & 22 & Binary, categorical, integer & Binary classification & 2 & 90.58\% / 9.42\% \\
\textit{Diabetes-H} & 1,000 (253,680) & 22 & Binary, categorical, integer & Multiclass classification & 3 & 84.24\% / 1.83\% / 13.93\% \\
\textit{Diabetes-130} & 1,000 (91,844) & 45 & Binary, categorical, integer & Multiclass classification & 3 & 53.67\% / 35.15\% / 11.18\% \\
\textit{Obesity} & 1,000 (2,111) & 17 & Binary, categorical, integer, continuous & Multiclass classification & 7 & Balanced classes (around 15\% each) \\
\textit{Obesity-Bin} & 1,000 (2,111) & 17 & Binary, categorical, integer, continuous & Binary classification & 2 & 54.10\% / 45.90\% \\
\textit{Breast Cancer} & 569 & 30 & Binary, categorical, continuous & Binary classification & 2 & 62.74\% / 37.26\% \\
\hline\hline
\end{tabular}
}
\caption{Summary of the clinical datasets used in this study. Subsets of 1,000 samples were extracted from the larger datasets to simulate realistic clinical scenarios, while the smallest dataset was used in its entirety.}
\label{tab:datasets}
\end{table*}

\subsection{Model Performance Across Clinical Datasets}\label{sec:ml_res}
To rigorously evaluate the predictive capabilities of our proposed models, \emph{Logistic-KAN} and \emph{KAAM}, we compare their performance against four widely used baselines: LR, Random Forest (RF), MLP, and NAM. All models are evaluated across the datasets described in Section \ref{sec:data}, under a consistent preprocessing and training pipeline. For each method, we perform a comprehensive hyperparameter optimization procedure as detailed in Appendix \ref{sec:app:hyperparam_opt}.

Table \ref{tab:res_ml} summarizes the results obtained for six datasets and five standard metrics: accuracy, ROC-AUC, F1-score, precision, and recall. Overall, both proposed models achieve highly competitive results across all tasks. Notably, \emph{Logistic-KAN} consistently obtains the highest recall and F1-score across multiple datasets (e.g., \textit{Heart}, \textit{Obesity}), reflecting its strong ability to capture class-specific patterns with balanced performance. \emph{KAAM}, in turn, demonstrates particularly robust results in terms of precision and AUC, even under its additive structure constraints. The inclusion of confidence intervals reinforces the reliability of these findings. In simpler binary tasks such as \textit{Heart} or Obesity-Bin, both \emph{Logistic-KAN} and \emph{KAAM} achieve near-perfect performance with very narrow intervals, suggesting high certainty in predictions. In more challenging settings, such as Diabetes-130, performance degrades as expected; however, both models remain competitive relative to ``black-box'' baselines. These results highlight the practical effectiveness of the proposed architectures in tabular clinical settings.

\renewcommand{\arraystretch}{1.35}
\begin{table*}[!ht]
    \centering
    \resizebox{\textwidth}{!}{%
    \begin{tabular}{l|l|cccccc}
    \hline\hline
    
     \multirow{2}{*}{\textbf{Dataset}}     & \multirow{2}{*}{\textbf{Model}}   & \multicolumn{5}{c}{\textbf{Classification Metrics}}   \\
     
    &   &   \textbf{Accuracy} &   \textbf{ROC-AUC} &    \textbf{F1-score} &   \textbf{Precision} &   \textbf{Recall} \\

    \hline\hline
     \multirow{6}{*}{\textit{Heart}}           & MLP          & \textbf{0.90 (0.85, 0.94)}   & 0.38 (0.25, 0.52) & 0.00 (0.00, 0.00)      & 0.00 (0.00, 0.00)      & 0.00 (0.00, 0.00)  \\
                & LR     & 0.79 (0.74, 0.85) & 0.89 (0.83, 0.95) & 0.44 (0.30, 0.58) & 0.31 (0.19, 0.44)  & 0.81 (0.63, 0.96)  \\
                & RF           & \textbf{0.90 (0.86, 0.94) } & 0.90 (0.83, 0.95)  & 0.54 (0.35, 0.71)  & \textbf{0.52 (0.31, 0.71)} & 0.57 (0.37, 0.79) \\
                & NAM          & \textbf{0.90 (0.85, 0.94) }  & 0.38 (0.26, 0.49)   & 0.00 (0.00, 0.00)      & 0.00 (0.00, 0.00)      & 0.00 (0.00, 0.00)  \\
                & Logistic-KAN          &       0.84 (0.79, 0.89)   & \textbf{0.91 (0.84, 0.96)} & \textbf{0.56 (0.41, 0.69)} & 0.40 (0.27, 0.53)  & \textbf{0.95 (0.85, 1.00)}   \\
                & KAAM      &      0.82 (0.77, 0.87)  & 0.90 (0.84, 0.95)  & 0.49 (0.34, 0.64) & 0.35 (0.22, 0.49) & 0.86 (0.69, 1.00) \\
          \hline
     \multirow{6}{*}{\textit{Diabetes-H}}           & MLP          & 0.81 (0.81, 0.82) & 0.58 (0.58, 0.59) & 0.75 (0.75, 0.75) & 0.71 (0.71, 0.72) & 0.81 (0.81, 0.82)  \\
           & LR         & 0.80 (0.80, 0.80)  & \textbf{0.70 (0.69, 0.70) }   & 0.81 (0.80, 0.81) & \textbf{0.81 (0.81, 0.82)} & 0.80 (0.80, 0.80)  \\
                & RF           &       \textbf{0.85 (0.85, 0.85)} & 0.66 (0.66, 0.67) & \textbf{0.82 (0.82, 0.82)} & \textbf{0.81 (0.81, 0.81) }& \textbf{0.85 (0.85, 0.85)}  \\
                & Logistic-KAN          &       0.82 (0.82, 0.86) & 0.68 (0.68, 0.69) & 0.75 (0.75, 0.76) & 0.73 (0.73, 0.74) & 0.82 (0.82, 0.83)\\
                & KAAM      &       0.82 (0.82, 0.83) & 0.68 (0.68, 0.69) & 0.75 (0.75, 0.76) & 0.73 (0.73, 0.74) & 0.82 (0.82, 0.83)  \\
     \hline
     
     \multirow{6}{*}{\textit{Diabetes-130}}           & MLP          &       0.52 (0.46, 0.59)  & \textbf{0.53 (0.46, 0.59) }& 0.45 (0.38, 0.53)   & 0.44 (0.36, 0.52) & 0.52 (0.46, 0.59)  \\
                & LR     &       \textbf{0.55 (0.48, 0.62)}  & \textbf{0.53 (0.46, 0.60)} & 0.50 (0.43, 0.58)   & 0.51 (0.43, 0.59) & \textbf{0.55 (0.48, 0.62)}  \\
                & RF           &       0.52 (0.46, 0.60) & \textbf{0.53 (0.46, 0.60)}   & \textbf{0.51 (0.44, 0.58)}  & 0.51 (0.44, 0.58) & 0.52 (0.46, 0.60) \\
                & Logistic-KAN          &       0.51 (0.44, 0.59)  & 0.49 (0.42, 0.57) & 0.46 (0.38, 0.54) & 0.46 (0.38, 0.54)  & 0.51 (0.44, 0.59)  \\
                & KAAM      &       \textbf{0.55 (0.49, 0.62) } &\textbf{ 0.53 (0.45, 0.60) } &\textbf{ 0.51 (0.43, 0.59)} & \textbf{0.55 (0.44, 0.63)} & \textbf{0.55 (0.49, 0.62) }\\
     
     \hline
     
\multirow{6}{*}{\textit{Obesity}}     & MLP          &       0.56 (0.50, 0.63)  & 0.88 (0.85, 0.90) & 0.55 (0.48, 0.62) & 0.56 (0.48, 0.64) & 0.56 (0.50, 0.63)  \\
          & LR     &       0.70 (0.64, 0.77)   & 0.92 (0.91, 0.94) & 0.68 (0.61, 0.75) & 0.68 (0.61, 0.75)  & 0.70 (0.64, 0.76)   \\
          & RF           &       0.91 (0.87, 0.95) & \textbf{1.00 (0.99, 1.00)}  & 0.91 (0.86, 0.95) & 0.92 (0.88, 0.95)  & 0.91 (0.87, 0.95)  \\
          & Logistic-KAN          &      \textbf{0.98 (0.95, 1.00)}  & \textbf{1.00 (1.00, 1.00) }   & \textbf{0.98 (0.95, 1.00)} & \textbf{0.98 (0.96, 1.00)} &\textbf{ 0.98 (0.95, 1.00)}  \\
          & KAAM      &       0.95 (0.92, 0.98)  & \textbf{1.00 (0.99, 1.00) }   & 0.95 (0.92, 0.98)  & 0.95 (0.92, 0.98)  & 0.95 (0.92, 0.98)  \\
     \hline
     
     \multirow{6}{*}{\textit{Obesity-Bin}} & MLP          &        0.86 (0.81, 0.91)  & 0.93 (0.90, 0.96) & 0.86 (0.80, 0.90) & 0.88 (0.81, 0.94) & 0.83 (0.76, 0.90)  \\
       & LR     &      \textbf{1.00 (0.99, 1.00) }   & \textbf{1.00 (1.00, 1.00)}      & \textbf{1.00 (0.98, 1.00)}    & 0.99 (0.97, 1.00)   & \textbf{1.00 (1.00, 1.00)}  \\
       & RF           &      0.99 (0.98, 1.00)   & \textbf{1.00 (1.00, 1.00)}    & 0.99 (0.97, 1.00)   & 0.99 (0.97, 1.00)   & 0.99 (0.97, 1.00)   \\
      & NAM          &       0.50 (0.43, 0.57)    & 0.89 (0.85, 0.93) & 0.00 (0.00, 0.00)      & 0.00 (0.0, 0.00)      & 0.00 (0.00, 0.00) \\
       & Logistic-KAN          &       \textbf{1.00 (1.00, 1.00)}      &\textbf{ 1.00 (1.00, 1.00)}      & \textbf{1.00 (1.00, 1.00)  }    &\textbf{ 1.00 (1.00, 1.00)}      & \textbf{1.00 (1.00, 1.00)}  \\
       & KAAM      &       \textbf{1.00 (1.00, 1.00)}      & \textbf{1.00 (1.00, 1.00)}      & \textbf{1.00 (1.00, 1.00)}      & \textbf{1.00 (1.00, 1.00)}      & \textbf{1.00 (1.00, 1.00)}  \\

     \hline
     
         \multirow{6}{*}{\textit{Breast Cancer}}      & MLP          &       0.94 (0.90, 0.98) & 0.98 (0.96, 1.00) & 0.95 (0.91, 0.99) & 0.92 (0.85, 0.98) & 0.98 (0.95, 1.00)   \\
          & LR     &       0.96 (0.91, 0.99) & \textbf{1.00 (0.99, 1.00)}    & 0.96 (0.92, 0.99) & \textbf{0.97 (0.92, 1.00)}   & 0.96 (0.90, 1.00)  \\
           & RF           &       0.95 (0.91, 0.99) & \textbf{1.00 (0.99, 1.00)}    & 0.96 (0.92, 0.99) & \textbf{0.97 (0.92, 1.00)}    & 0.95 (0.90, 1.00)  \\
           & NAM          &       0.92 (0.87, 0.97) & 0.98 (0.96, 1.00) & 0.93 (0.89, 0.97) & 0.91 (0.84, 0.97) & 0.96 (0.90, 1.00) \\
           & Logistic-KAN          &       \textbf{0.97 (0.94, 1.00)}   & 0.99 (0.98, 1.00)    & \textbf{0.98 (0.95, 1.00) }  & \textbf{0.97 (0.92, 1.00)}   & \textbf{0.99 (0.95, 1.00)} \\
           & KAAM      &       0.96 (0.93, 0.99)  & \textbf{1.00 (0.99, 1.00) }   & 0.97 (0.94, 0.99) & \textbf{0.97 (0.93, 1.00)}   & 0.97 (0.92, 1.00) \\

    \hline\hline
    \end{tabular}}
    
    \caption{Classification performance of the evaluated models across six clinical datasets. Each value is reported with its associated 95\% confidence interval, computed via bootstrapping. The best mean performance for each metric and dataset is highlighted in \textbf{bold}. Higher values indicate better performance. Note that NAM only supports binary classification.}
    \label{tab:res_ml}
\end{table*}

\begin{figure}[!ht]
\centering
\begin{minipage}[c]{0.42\linewidth}
    \centering
    \renewcommand{\arraystretch}{1.35}
    \resizebox{\linewidth}{!}{%
    \begin{tabular}{l|cccccc}
    \hline\hline
    \textbf{Classification} & \multicolumn{6}{c}{\textbf{Model}} \\
    \textbf{Metric} & \textbf{MLP} & \textbf{LR} & \textbf{RF} & \textbf{NAM} & \textbf{Logistic-KAN} & \textbf{KAAM} \\
    \hline\hline
    \textit{Accuracy} & 0.44 & 0.54 & 0.61 & 0.48 & \textbf{0.72} & 0.64\\
    \textit{ROC-AUC}  & 0.44 & 0.81 & 0.81 & 0.31 & 0.75 & \textbf{0.83} \\
    \textit{F1-score} & 0.26 & 0.47 & 0.61 & 0.21 & \textbf{0.78} & 0.61 \\
    \textit{Precision}& 0.30 & 0.58 & \textbf{0.72} & 0.26 & \textbf{0.72} & \textbf{0.72} \\
    \textit{Recall}   & 0.34 & 0.51 & 0.46 & 0.23 & \textbf{0.81} & 0.64 \\
    \hline
    \textbf{\textit{All metrics}} & 0.36 & 0.58 & 0.64 & 0.30 & \textbf{0.76} & 0.69 \\
    \hline\hline
    \end{tabular}
    }
    \captionof{table}{MRR for each model across datasets. \textbf{Bold} represents the best MRR values for each metric. Higher is better.}
    \label{tab:res_mrr}
\end{minipage} \hfill
\begin{minipage}[c]{0.55\linewidth}
    \centering
    \includegraphics[width=\linewidth]{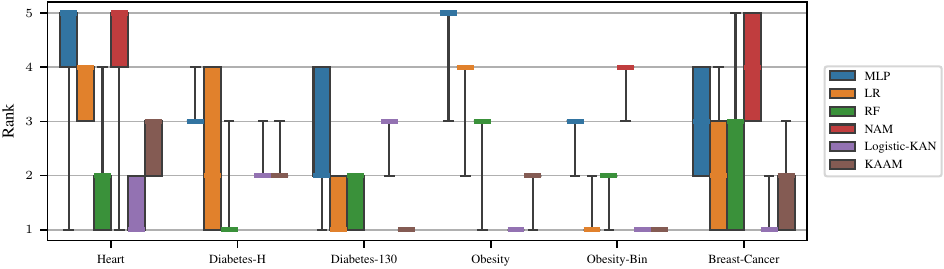}
    \caption{Boxplots of rankings, aggregating all the classification metrics, across datasets and models. Collapsed boxplots indicate a model that ranks the same across all metrics. Lower is better.}
    \label{fig:ranking_boxplots}
\end{minipage}
\end{figure}

To aggregate performance across datasets and metrics in a principled way, we compute the Mean Reciprocal Rank (MRR). For each experiment, the candidate models are ranked according to their metric value, and the reciprocal of their rank is averaged across tasks:

\begin{equation*}
    \mathrm{MRR} = \frac{1}{N} \sum_{i=1}^{N} \frac{1}{r_i},
\end{equation*}

where $r_i$ denotes the rank of a model in task $i$, and $N$ is the number of evaluated tasks. The MRR values per metric and the global MRR across all metrics are reported in \cref{tab:res_mrr}. 

\emph{Logistic-KAN} achieves the highest overall MRR ($0.76$), outperforming all baselines. It leads in accuracy, F1-score, precision, and recall, demonstrating both strong predictive performance and robustness across datasets. \emph{KAAM} follows closely with an overall MRR of $0.69$, outperforming all other models in ROC-AUC and precision, and showing competitive results in the remaining metrics. In contrast, LR and RF perform well but lag behind the proposed models, particularly in terms of accuracy and recall. Neural models such as MLP and NAM exhibit substantially lower MRRs, highlighting their weaker consistency across clinical datasets in this setting. While this may partially reflect architectural limitations, such as a known tendency to overfit or difficulties in modeling categorical variables, it may also be influenced by factors such as feature encoding or dataset size. An additional representation that helps to understand the performance of the models across metrics can be found in \cref{fig:ranking_boxplots}, where we can see that there is at least one of the proposed models (\emph{Logistic-KAN} or KAAM) among the best performers in all datasets, aggregating by all classification metrics.

To complement these aggregate rankings with formal statistical evidence, we provide a significance analysis based on the Friedman test and post-hoc pairwise comparisons in Appendix \ref{sec:app:stats} \citep{demvsar2006statistical, rainio2024evaluation}. This complementary perspective confirms that the observed improvements, particularly those of \emph{Logistic-KAN} and \emph{KAAM}, are statistically robust across multiple datasets and metrics.

These findings validate the effectiveness of our approach in terms of pure predictive performance. Importantly, this is achieved without compromising interpretability, unlike conventional neural models, thus offering a compelling trade-off for clinical applications. In the following sections, we further demonstrate how \emph{KAAM} supports interpretations through tailored visualization tools and individual-level analysis.

\subsection{Visual Interpretability}\label{sec:xai_res}
To showcase the built-in interpretability capabilities of \emph{KAAMs}, we analyze the model’s behavior on the Diabetes-130 dataset, a multiclass diagnostic classification task. Unlike traditional ``black-box'' models that require external tools to explain their outputs, \emph{KAAMs} offer transparency by design. This makes them especially well-suited for clinical applications, where understanding the reasoning behind a prediction is often as important as the prediction itself.

\emph{KAAM} enables two complementary types of interpretability: global (population-level) interpretations, which help clinicians understand which features drive the model’s decisions across the entire cohort, and individual (patient-level) interpretations, which reveal how specific covariates influence the prediction for a given patient. This dual perspective is essential for precision medicine, empowering clinicians to understand not only what the model predicts but also why (both at the cohort and individual levels). 

We focus on \emph{KAAM} for this analysis as its additive structure allows a straightforward decomposition of each prediction into feature-level contributions. While the full technical description is deferred to Section \ref{sec:methods}, it is worth recalling that \emph{KAAM’s} structural constraints result in more compact symbolic formulas and visually intuitive outputs, making it particularly well-suited for clinically actionable interpretability. In Appendix \ref{sec:app:more_results}, we show the full analytical expression for the \emph{KAAM} model trained on this dataset. Although the resulting function is structurally complex, its interpretability is enhanced through visualization techniques that facilitate the understanding of the model's reasoning. 

We demonstrate this interpretability using three complementary visual tools: (i) PDPs to assess individual feature contributions; (ii) feature importance analysis to understand global model behavior; and (iii) PRPs to compare patient-specific profiles with the cohort average. These are further complemented by nearest-patient comparisons and interactive \textit{model-response} simulations, offering a multifaceted view of model predictions.

\begin{figure*}[!ht]
\centering
    \begin{subfigure}{0.32\linewidth}
        \includegraphics[width=\linewidth]{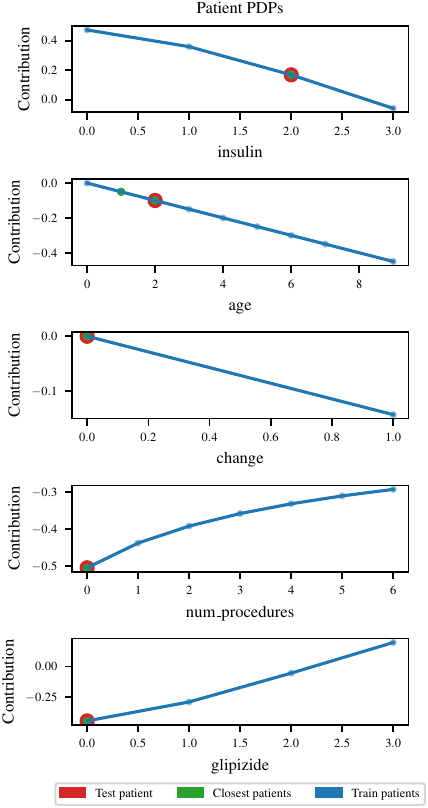}
        \caption{Class $0$}
    \end{subfigure}
    \begin{subfigure}{0.32\linewidth}
        \includegraphics[width=\linewidth]{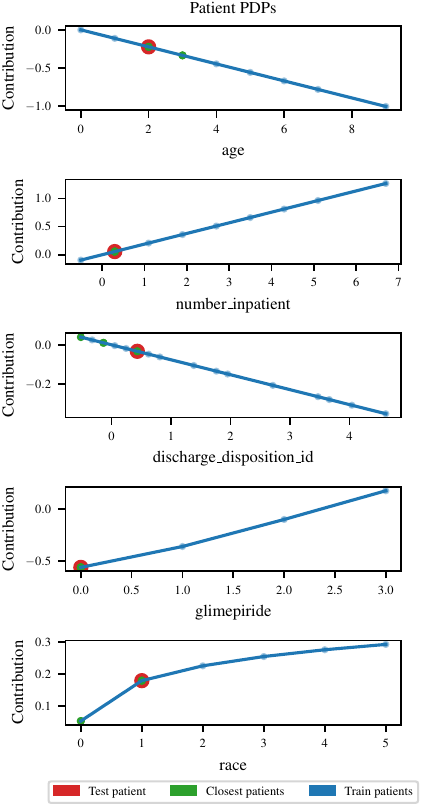}
        \caption{Class $1$}
    \end{subfigure}
    \begin{subfigure}{0.32\linewidth}
        \includegraphics[width=\linewidth]{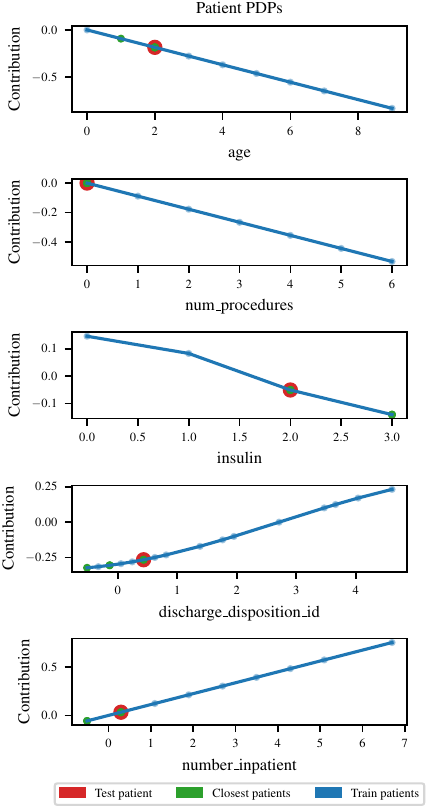}
        \caption{Class $2$}
    \end{subfigure}
    \caption{PDPs for a test patient in the Diabetes-130 dataset using the \emph{KAAM} model. Each subplot illustrates how individual covariates affect the predicted probability for a specific class. Blue lines represent the \emph{KAAM}-learned covariate contribution, orange dots mark the current patient's values, light blue dots correspond to the training cohort, and green dots show the nearest neighbors.}
    \label{fig:pdd_4}
\end{figure*}

\paragraph{Partial Dependence Plots.}
\emph{KAAM} enables the computation of PDPs, which visualize how the predicted probability of a given class changes as one covariate is varied, while the others are kept fixed. These plots help clinicians understand not only whether a covariate increases or decreases risk, but also how sensitive the model is to changes in its value. The blue curve shows the contribution of each covariate to the predicted probability for each class. The red dot indicates the covariate value for the patient under analysis, while green and light blue dots correspond to the most similar patients and the training cohort, respectively. For instance, in the leftmost panel (Class 0) in Figure \ref{fig:pdd_4}, we observe that higher values of \textit{Glipizide} are associated with increased risk, whereas \textit{Insulin} appears to reduce the predicted probability. These plots are especially helpful for supporting counterfactual reasoning and personalized clinical insight \citep{loftus2024causal}. Note that the term “counterfactual” is used here in the sense of counterfactual explanations in machine learning \citep{wachter2017counterfactual}, namely as hypothetical input modifications that alter the model’s prediction. This notion differs from causal counterfactuals, which require a structural causal analysis grounded in an explicit causal graph and assumptions about observed and unobserved variables, including potential sources of bias such as confounders, colliders, and mediators \citep{pearl2009causality}.

\paragraph{Feature Importance Analysis.}
\emph{KAAM} naturally provides a global measure of feature importance, allowing us to rank covariates by their contribution to the model’s predictions at the population level. This information is essential for clinicians to understand which variables are driving the diagnostic decisions across the entire cohort. 

\begin{figure*}[!ht]
    \centering
    \begin{subfigure}{0.32\linewidth}
        \includegraphics[width=\linewidth]{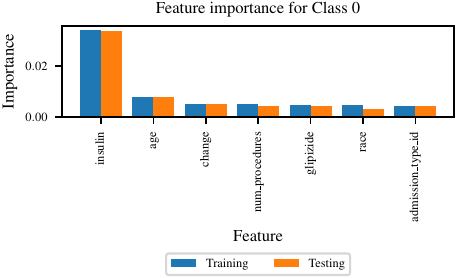}
        \caption{Class $0$}
    \end{subfigure}
    \begin{subfigure}{0.32\linewidth}
        \includegraphics[width=\linewidth]{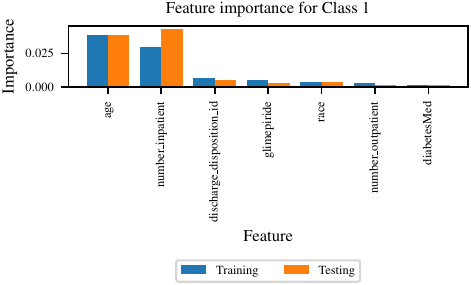}
        \caption{Class $1$}
    \end{subfigure}
    \begin{subfigure}{0.32\linewidth}
        \includegraphics[width=\linewidth]{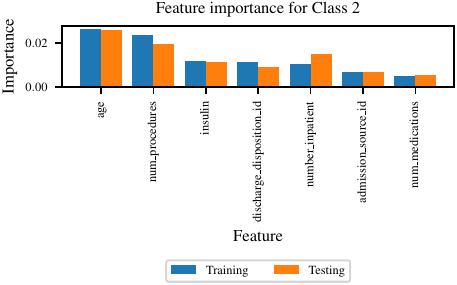}
        \caption{Class $2$}
    \end{subfigure}
    \caption{Importance plot for the patients in the Diabetes-130 dataset using the \emph{KAAM} model. Not all covariates are equally important for predicting each class, although some covariates (\textit{Age} or \textit{Insulin}) are important across classes. Deviations between train and test patients are minimal.}
    \label{fig:var_4}
\end{figure*}

Figure \ref{fig:var_4} illustrates this for the Diabetes-130 dataset. For each diagnostic class, the plot shows which features contribute most to the corresponding class function. For example, in Class 0, the most influential covariates are \textit{Insulin} and \textit{Age}, both of which are clinically meaningful. Interestingly, \textit{Age} is consistently among the most relevant variables across all three classes. Moreover, the similarity in importance rankings between the training and test patients suggests that the model generalizes well and that its reasoning remains stable across different subgroups.

\paragraph{Probability Radar Plots.}
PRPs provide a compact and intuitive graphical summary of which covariates are contributing to a patient’s predicted probability being above or below the population average. Each axis corresponds to a single input covariate, and the plotted value reflects the model's predicted probability for that class when that covariate is fixed to the patient’s actual value. At the same time, all other features are set to their population average.

\begin{figure*}[!ht]
\centering
    \begin{subfigure}{0.32\linewidth}
        \includegraphics[width=\linewidth]{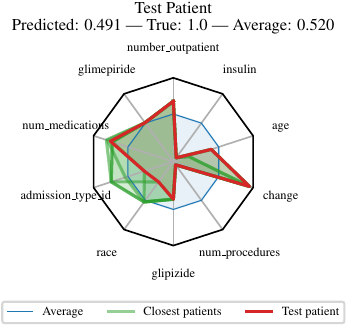}
        \caption{Class $0$}
    \end{subfigure}
    \begin{subfigure}{0.32\linewidth}
        \includegraphics[width=\linewidth]{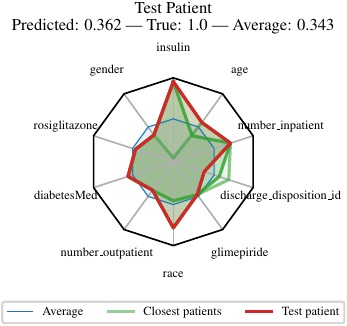}
        \caption{Class $1$}
    \end{subfigure}
    \begin{subfigure}{0.32\linewidth}
        \includegraphics[width=\linewidth]{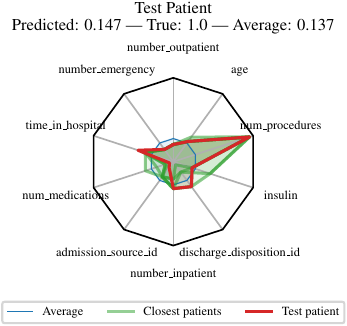}
        \caption{Class $2$}
    \end{subfigure}

    \caption{PRPs for a representative test patient from the Diabetes-130 dataset, showing the individual contribution of each covariate to the predicted class probabilities. Each axis corresponds to a covariate. The orange polygon represents the class probability when the covariate is fixed to the patient's actual value and all other features are set to their population averages. The blue polygon represents the average patient (mean value for all features), and the green polygon corresponds to the closest training sample. In Class 2, for example, the patient’s \textit{num\_procedures} significantly increases the risk, while \textit{num\_medications} decreases it. The similarity in polygon shapes between the orange and green curves provides an intuitive indication of how familiar or atypical a patient is compared to historical data.}
    
    \label{fig:radar_4}
\end{figure*}

This visualization enables clinicians to instantly identify risk-driving variables: covariates whose values push the orange polygon outside the blue region indicate an increased risk compared to the average patient; values inside indicate a protective effect. For instance, in the PRP for Class 0, the patient’s \textit{Insulin} level lowers the predicted risk compared to the cohort mean. In contrast, their \textit{Age} places them near the population baseline. In addition to the average (blue) and patient-specific (orange) predictions, the PRP also includes a green polygon representing the nearest patient in the model’s internal representation space. When the orange and green shapes closely align, it suggests that the model is basing its prediction on well-established patterns from similar cases. If the orange shape deviates significantly from the green, it may indicate a less typical or outlier case that warrants special attention (see Figure \ref{fig:radar_4}).

\paragraph{Nearest Patients Retrieval}
Beyond visual interpretation, \emph{KAAMs} provide a powerful capability for case-based reasoning by identifying the most similar patients from the training set. This retrieval mechanism functions as a clinical ``search engine,'' enabling clinicians to locate previous cases with highly similar profiles to the patient being evaluated.

Accessing these neighboring patients allows the practitioner to examine past outcomes, diagnoses, and treatment decisions, providing real-world context for the current prediction. For example, suppose the retrieved patients (those with risk profiles similar to the current one) did not develop critical conditions. In that case, this may support a more conservative decision. Conversely, if those patients experienced adverse events, this could justify earlier intervention for the current patient.

Importantly, similarity is computed in the \emph{KAAM} logit space, which captures the model’s internal representation of risk rather than relying on raw feature proximity. We use the Euclidean distance between the patient's logit vector and those of the training samples to retrieve clinically relevant cases aligned with the model’s own reasoning. This offers a more clinically meaningful notion of similarity rooted in the predictive behavior of the model.

\paragraph{Interactive Interfaces and Model-Response Analysis}
The symbolic and interpretable nature of \emph{KAAMs} naturally lends itself to the development of interactive tools for clinical use. We envision these visualizations being integrated into a decision-support interface that allows real-time inspection and simulation of patient scenarios.

In particular, clinicians can conduct ``what-if'' analysis by modifying individual covariates and observing how this affects key interpretability plots, such as PDPs, PRPs, and the retrieved nearest patients \footnote{Note that this ``what-if'' concept refers to prediction changes with input perturbations, not to causal hypothetical interventions in the real-world. In a causal analysis, a change in one variable may affect others, depending on the causal graph.}. This supports the dynamic exploration of clinical questions, such as: \emph{What does my model predict if this patient did not smoke, keeping the rest variables constant?} Or, \emph{how would the risk profile change if their blood pressure had been higher?}

Such capabilities promote transparency and empower healthcare professionals to incorporate domain knowledge directly into the decision-making process. An example of a working prototype is available in our public code repository (\url{https://github.com/Patricia-A-Apellaniz/classification_with_kans/src/use_case.ipynb}), built using the \textit{Heart} dataset. This prototype addresses a binary classification task to predict the presence or absence of heart disease. Clinicians can input new patient profiles, modify feature values, and observe changes in real-time across different interpretability visualizations.

\subsection{Use Case: Symbolic and Personalized Interpretations on the Heart Dataset} \label{sec:usecase}

To illustrate the practical value of \emph{KAAM’s} built-in interpretability for clinical applications, we present a guided use case using the Heart dataset. This binary classification task involves predicting whether a patient has suffered from heart disease or a heart attack. The dataset is notably imbalanced, with only 10\% of patients belonging to the positive class (i.e., those who have experienced a cardiovascular event).

We train our \emph{KAAM} model on the dataset, which in this case learns the following analytical approximation for the logit $l(x_i)$ of the positive class:

\begin{equation}
    \label{eq:formula_7}
    \begin{split}
        l =& \quad0.51 \times \mathrm{DiffWalk} + 0.49 \times \mathrm{HighChol}
        + 1.4 \times \mathrm{Sex} + 0.21 \times \mathrm{Smoker} \\
        \quad&+ 0.82 \times \mathrm{Stroke} + 1.14 \times \sin{\left(0.79 \times \mathrm{Age} + 0.6 \right)} - 1.08 \times \sin{\left(0.8 \times \mathrm{Age} - 2.60 \right)} \\
        \quad&+ 0.13 \times \cos{\left(1.43 \times \mathrm{PhysHlth} - 7.44 \right)} + 1.17 \times \tanh{\left(0.63 \times \mathrm{GenHlth} + 0.91 \right)} \\
        \quad&+ 0.74 \times \tanh{\left(0.71 \times \mathrm{GenHlth} + 0.86 \right)} - 4.27.
    \end{split}
\end{equation}

The predicted probability is obtained via $P(y_i = 1 \mid x_i) = \sigma(l(x_i))$, where $\sigma$ denotes the sigmoid function. As seen in \Eqref{eq:formula_7}, the \emph{KAAM} model has learned a mixture of linear and nonlinear effects, capturing linear contributions for binary features (e.g., \textit{Smoker}, \textit{Stroke}) and more complex behaviors for continuous variables (e.g., \textit{Age}, \textit{PhysHlth}, \textit{GenHlth}).

Before exploring interpretability tools, it is crucial to verify that the model produces reliable predictions. Table \ref{tab:ml_metrics_7} reports the classification performance of \emph{KAAM} under three conditions: (1) directly using the learned B-splines (i.e., smooth basis functions used to approximate nonlinear relationships in KAN architectures), (2) replacing them with symbolic expressions, and (3) rounding the coefficients in the symbolic expression to three decimal places, as shown in \Eqref{eq:formula_7}. Interestingly, the rounded symbolic version slightly outperforms the others across all metrics. While these differences are not statistically significant, they suggest that the symbolic approximation not only preserves but may even enhance generalization by acting as a mild regularizer. Most importantly, the results confirm that interpretability can be achieved without sacrificing predictive performance, validating the robustness of \emph{KAAM’s} symbolic representations.

\renewcommand{\arraystretch}{1.35}
\begin{table}[!ht]
    \centering
    \resizebox{0.6\textwidth}{!}{%
    \begin{tabular}{l|ccc}
    \hline\hline
          \textbf{Metric}    & \textbf{Test}     & \textbf{Formula}  & \textbf{Formula + 3 decimals}    \\ \hline\hline
    \textit{Accuracy}  & 0.824 (0.823,0.826)  & 0.825 (0.823,0.827) &  \textbf{0.851 (0.849,0.852)} \\
    \textit{AUC}       & \textbf{0.914 (0.912,0.915)} &\textbf{ 0.911 (0.909,0.913)} &  \textbf{0.913 (0.911,0.916)} \\ 
    \textit{F1-score}        & 0.515 (0.511,0.520) & 0.517 (0.512,0.521) &  \textbf{0.558 (0.553,0.562)} \\
    \textit{Precision} & 0.363 (0.359,0.367)   & 0.366 (0.361,0.370) &  \textbf{0.407 (0.402,0.411)} \\ 
    \textit{Recall}    & \textbf{0.907 (0.903,0.911)}  & \textbf{0.902 (0.898,0.907)}   & \textbf{0.906 (0.901,0.910) }\\ \hline \hline
    \end{tabular}}
    \caption{Performance of the \emph{KAAM} model on the Heart dataset under different levels of symbolic approximation. Each value is reported alongside its associated 95\% confidence interval, computed via bootstrapping. Higher is better for all metrics. \textbf{Bold} represents the best values for each metric.}
    \label{tab:ml_metrics_7}
\end{table}

\begin{figure}[!ht]
\centering
    \begin{subfigure}{0.45\linewidth}
        \includegraphics[width=\linewidth]{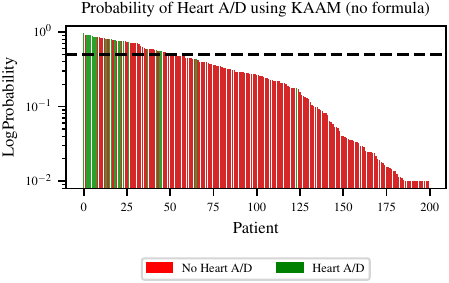}
        \caption{Predictions with B-splines}
    \end{subfigure}
    \begin{subfigure}{0.45\linewidth}
        \includegraphics[width=\linewidth]{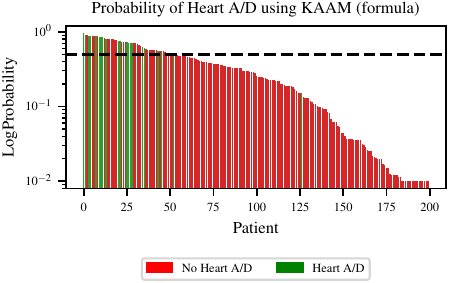}
        \caption{Predictions with symbolic formula}
    \end{subfigure}
    \caption{Predicted probabilities for test patients in the Heart dataset using \emph{KAAM}. Each bar represents one patient, colored by true class (red: negative, green: positive). Predictions are shown using the original B-spline model (left) and its symbolic approximation (right). The dashed line denotes the classification threshold (0.5).}
    \label{fig:ml_plots_7}
\end{figure}

Figure \ref{fig:ml_plots_7} visually compares the predicted probabilities for test patients under both the spline-based and symbolic \emph{KAAM} representations. Patients are sorted along the horizontal axis, and the height of each bar indicates the predicted probability of belonging to the positive class. The color encodes the true class label (red: negative, green: positive), while the dashed line denotes the decision threshold (default: 0.5). This visualization confirms the model's ability to detect high-risk individuals (few false negatives), aligning with the high recall in Table \ref{tab:ml_metrics_7}. At the same time, the model also predicts a substantial number of false positives (red bars above the threshold), reflecting a common trade-off between sensitivity and specificity. Importantly, this figure offers complementary insight to threshold-independent metrics such as ROC-AUC. While AUC provides a summary of the model's overall discriminative power, it does not reveal how individual predictions are distributed relative to the decision boundary. In clinical settings, this level of detail is often crucial: for example, reducing the number of false negatives may be more desirable than avoiding false positives. In such cases, a clinician may prefer to lower the decision threshold to ensure that no high-risk patients are missed, even if this leads to more false alarms. This visualization explicitly supports such reasoning, making it a valuable tool for interpretability in high-stakes applications, such as cardiovascular risk prediction.

\begin{figure}[!ht]
    \centering
    \begin{subfigure}{0.45\linewidth}
        \includegraphics[width=\linewidth]{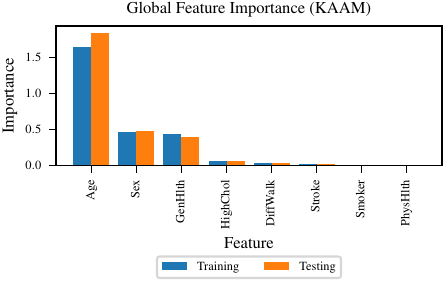}
        \caption{Importance predicted with \emph{KAAM}}
    \end{subfigure}
    \begin{subfigure}{0.45\linewidth}
    \includegraphics[width=\linewidth]{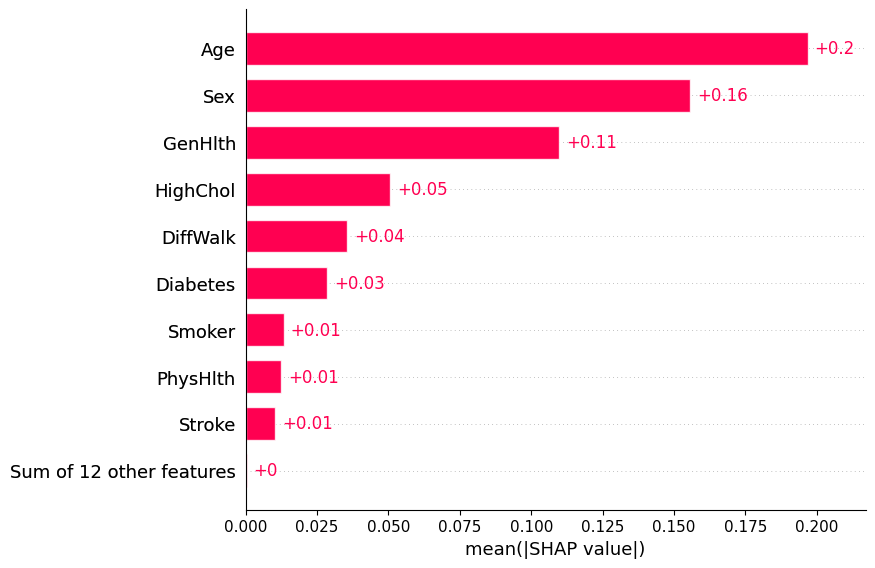}
        \caption{Importance predicted with SHAP}
    \end{subfigure}
    \caption{Importance of the covariates in the Heart dataset predicted using the \emph{KAAM} variance and SHAP. Both methods yield very similar results, emphasizing the same features, which confirms the correctness of the proposed method and aligns with clinical intuition (i.e., \emph{Age} is a good predictor of heart disease).}
    \label{fig:imp_7}
\end{figure}

In addition to predictive accuracy, a key aspect for clinical deployment is the ability to identify which covariates contribute most to the model's decision. In other words, clinicians need to understand which patient variables are most predictive of risk. \emph{KAAMs} provide this capability natively. Figure \ref{fig:imp_7} compares the global feature importance scores obtained from \emph{KAAM} with those derived using SHAP, a widely adopted post-hoc explanation method. Importantly, SHAP is computed over the same trained \emph{KAAM} model, using its prediction function as input to the SHAP explainer. The results exhibit strong agreement between both techniques, consistently highlighting the same variables, such as \textit{Age}, \textit{Sex}, and \textit{GenHlth}, as the most relevant predictors. This alignment further validates the trustworthiness of the \emph{KAAM} importance scores. However, \emph{KAAM} offers two notable advantages over SHAP. First, it is significantly more computationally efficient: while SHAP must be computed after training the predictive model (often requiring sampling and retraining), the \emph{KAAM} importance scores follow directly from model training. Second, \emph{KAAM} provides inherent importance scores, rather than post-hoc approximations, avoiding potential distortions introduced during model explanation. One potential caveat of variance-based importance metrics is their sensitivity to feature scaling. However, this effect is mitigated in our experiments through standard preprocessing techniques (e.g., feature normalization), ensuring comparability across covariates.

\begin{figure}[!h]
\centering
    \begin{subfigure}{0.55\linewidth}
        \includegraphics[width=.8\linewidth]{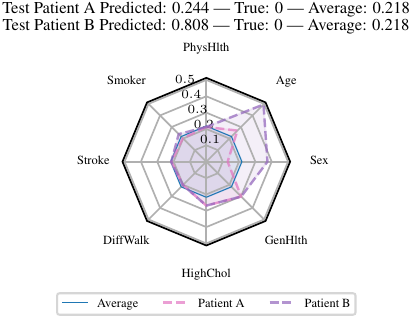}
        \caption{PRP}
    \end{subfigure}
    \begin{subfigure}{0.4\linewidth}
        \includegraphics[width=\linewidth, trim={0, 0, 0, 0}, clip]{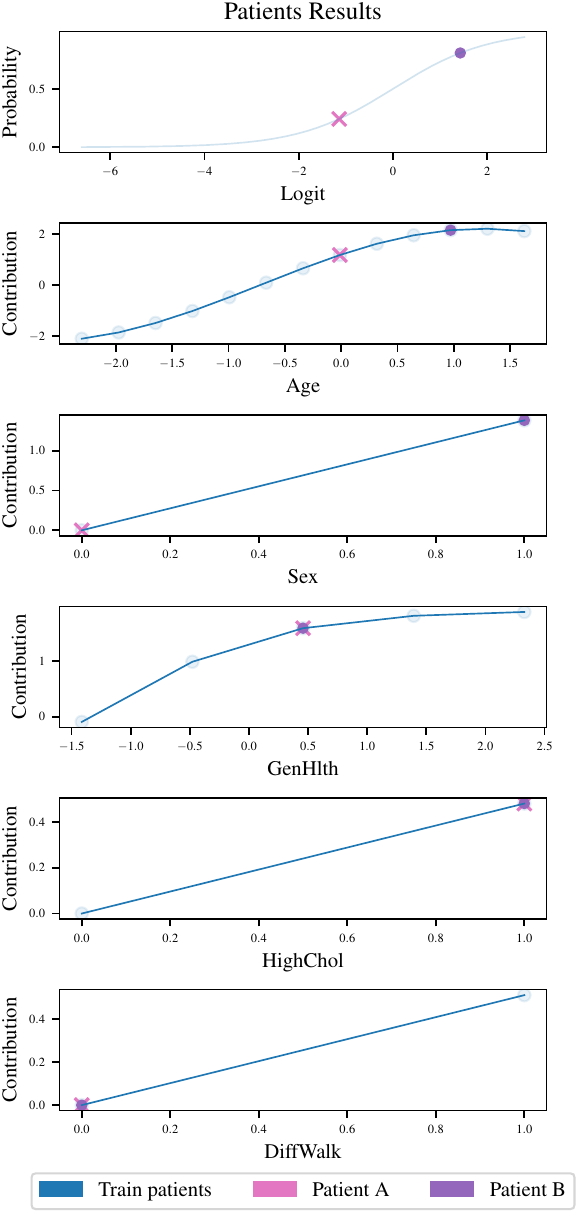}
        \caption{PDP}
    \end{subfigure}
        
     \caption{PRP and PDP for two test patients (A in pink, B in purple) in the Heart dataset. The PRP subfigure (a) plot shows how the risk changes as the covariate varies. Colored markers indicate the value of each patient, enabling clinicians to identify which features contribute more or less to the predicted risk. The PDP subfigure (b) shows how each covariate influences the class prediction, highlighting the personalized risk profile of each patient. The blue line represents the univariate contribution function in the range of each variable. The overlaid pink and purple points show the individual values for each patient, allowing a direct comparison of their risk profiles.}
     \label{fig:pdd_prp_7}
\end{figure}

Beyond global model performance and feature importance, one of the most valuable capabilities of \emph{KAAMs} lies in their capacity to provide personalized, patient-specific insights. When a new patient arrives, the \emph{KAAM} not only produces a probabilistic prediction (e.g., likelihood of developing a heart condition) but also delivers an individualized interpretation grounded in the patient’s own covariates. These interpretations can be visualized using tools such as PDPs and PRPs. To illustrate this, we consider two patients from the test set of the Heart dataset, denoted as Patient A and Patient B. Patient A is predicted to have a low risk of heart disease ($P(y_i=1)=0.244$), and indeed is a true negative. Patient B, on the other hand, is assigned a high risk ($P(y_i = 1) = 0.808$), but in reality, does not suffer from the condition, hence, a false positive. Figure \ref{fig:pdd_prp_7} displays the \emph{KAAM}-derived interpretations for two patients. The PRP summarizes the marginal contribution of each covariate to the patients' predicted risk, compared to the average patient. For instance, both patients exhibit high-risk values for variables such as \textit{Age}, \textit{GenHlth}, and \textit{HighChol}. However, Patient B presents additional risk-enhancing factors, such as being male and a smoker. Conversely, Patient A benefits from protective covariate values that counterbalance their risk. This visual summary enables clinicians to quickly identify which covariates are driving the model’s prediction and to what extent. To further investigate how each covariate affects the predicted outcome, the PDP offers a more detailed perspective. For example, the learned function for \textit{Age} exhibits a nonlinear increase in risk beyond a certain threshold, while variables such as \textit{Stroke} contribute positively to the risk, in line with clinical expectations. These plots also indicate the model's sensitivity to local changes in each feature, making them especially useful for counterfactual reasoning\footnote{In the sense of counterfactual explanations, see PDP details in \cref{sec:xai_res}.}.

In addition to explaining predictions, \emph{KAAMs} enable case-based reasoning by retrieving the most similar patients from the training set. This provides a form of clinical memory, allowing the practitioner to examine outcomes and treatments for similar cases. Tables \ref{tab:closest_7_A} and \ref{tab:closest_7_B} show the five closest patients for Patients A and B, respectively. Patient A’s nearest neighbors are all true negatives, reinforcing confidence in the model’s prediction. In contrast, most of Patient B’s nearest neighbors are also false positives, signaling that the model tends to overestimate risk in this region of the feature space. This information can guide the clinician in overriding or contextualizing the prediction based on prior patient outcomes.

\renewcommand{\arraystretch}{1.35}
\begin{table*}[!ht]
    \centering
    \resizebox{0.7\textwidth}{!}{%
    \begin{tabular}{cccccccc|c|c}
    \hline\hline
    \textbf{Age}    & \textbf{Sex} & \textbf{GenHlth} & \textbf{HighChol} & \textbf{DiffWalk} & \textbf{Stroke} & \textbf{Smoker} & \textbf{PhysHlth} & $\bm{P(y_i=1)}$ & \textbf{Real label} \\ \hline\hline
     -0.011 & 0.0 & 0.457   & 1.0      & 0.0      & 0.0    & 0.0    & -0.487   & 0.250      & 0.0         \\ \hdashline
     -0.011 & 0.0 & 0.457   & 1.0      & 0.0      & 0.0    & 0.0    & -0.028   & 0.251      & 0.0         \\ 
     -0.011 & 0.0 & 0.457   & 1.0      & 0.0      & 0.0    & 1.0    & -0.487   & 0.291      & 0.0         \\ 
     -0.011 & 0.0 & 0.457   & 1.0      & 0.0      & 0.0    & 1.0    & -0.372   & 0.290      & 0.0         \\ 
     -0.011 & 0.0 & 0.457   & 1.0      & 0.0      & 0.0    & 1.0    & 0.087    & 0.294      & 0.0         \\ 
     -0.011 & 0.0 & 0.457   & 1.0      & 0.0      & 0.0    & 1.0    & 2.955    & 0.320      & 0.0         \\ \hline\hline
    \end{tabular}}
        \caption{Covariate values of Patient A (first row) and the five closest training samples. The majority of these neighbors are false positives, indicating a local pattern of overestimation by the model. This observation can help the clinician reassess the prediction and identify potential model limitations in this region of the feature space.}
    \label{tab:closest_7_A}
\end{table*}

\renewcommand{\arraystretch}{1.35}
\begin{table*}[!ht]
    \centering
    \resizebox{0.7\textwidth}{!}{%
    \begin{tabular}{cccccccc|c|c}
    \hline\hline
    \textbf{Age}    & \textbf{Sex} & \textbf{GenHlth} & \textbf{HighChol} & \textbf{DiffWalk} & \textbf{Stroke} & \textbf{Smoker} & \textbf{PhysHlth} & $\bm{P(y_i=1)}$ & \textbf{Real label} \\ \hline\hline
     0.972 & 1.0 & 0.457   & 1.0      & 0.0      & 0.0    & 1.0    & -0.487   & 0.813      & 0.0         \\ \hdashline
     0.972 & 1.0 & 0.457   & 1.0      & 0.0      & 0.0    & 1.0    & -0.487   & 0.813      & 0.0         \\ 
     0.972 & 1.0 & 0.457   & 1.0      & 0.0      & 0.0    & 1.0    & -0.142   & 0.813      & 1.0         \\ 
     0.972 & 1.0 & 0.457   & 1.0      & 0.0      & 0.0    & 1.0    & -0.028   & 0.814      & 0.0         \\ 
    1.299 & 1.0 & 0.457   & 1.0      & 0.0      & 0.0    & 1.0    & -0.487   & 0.821      & 0.0         \\ 
     1.299 & 1.0 & 0.457   & 1.0      & 0.0      & 0.0    & 1.0    & -0.487   & 0.821      & 0.0         \\ \hline\hline
    \end{tabular}}
    \caption{Covariate values of Patient B (first row) and the five closest training samples. In this case, most of the closest patients to B are false positives, so this alerts the clinician that patient B may be a false positive (which is indeed the case).}
    \label{tab:closest_7_B}
\end{table*}

\begin{figure*}[!ht]
\centering
   \includegraphics[width=0.3\linewidth]{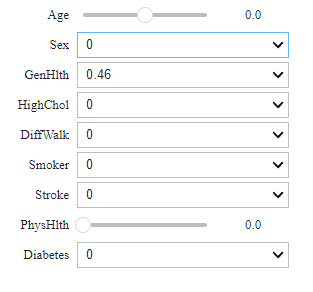}
    \caption{Example of an interactive interface for patient assessment in the Heart dataset. Clinicians can input patient covariates and explore the resulting interpretation tools, including the PDP and PRP (Figure \ref{fig:pdd_prp_7}), as well as the list of closest patients in the logit space (Tables \ref{tab:closest_7_A} and \ref{tab:closest_7_B}). This interface supports counterfactual exploration and is available in our public code repository.}
    \label{fig:interface_7}
\end{figure*}

Finally, these interpretability tools can be integrated into interactive platforms that allow clinicians to perform \textit{input perturbation} analyses, such as assessing how modifying a given covariate (e.g., \emph{Smoker} or \emph{Age}) would affect the predicted risk. An example of such an interface is shown in Figure \ref{fig:interface_7}, implemented as a Jupyter Notebook available in our public repository at \url{https://github.com/Patricia-A-Apellaniz/classification_with_kans}. This interface enables data entry, generates explanatory plots, and retrieves similar patients, offering a transparent and practical decision support tool for clinical environments.

%% file: 4_discussion.tex
\section{Discussion}
\label{sec:discussion}

In this work, we explored KANs as a powerful alternative to ``black-box'' neural models for medical classification tasks. We introduced two novel architectures: the \emph{Logistic-KAN}, which generalizes classical LR through flexible symbolic parameterizations, and the \emph{KAAM}, a simplified yet expressive variant designed for transparent decision-making.

Our experimental results across six clinical datasets highlight that both models achieve competitive classification performance when compared to standard baselines, including deep learning methods and NAMs. While \emph{Logistic-KAN} offers greater modeling capacity due to its compositional structure, we found that \emph{KAAM} consistently delivers robust, interpretable results, even in settings with limited data or imbalanced classes. This is particularly relevant in clinical contexts, where data may be scarce, and decisions must be interpretable.

What sets \emph{KAAM} apart is its built-in interpretability, designed by default. Unlike post-hoc methods such as SHAP or LIME, which approximate the model's behavior externally, \emph{KAAMs} generate interpretations intrinsically. Thanks to the analytical logit function learned during training, clinicians can directly access a symbolic formula that governs the model's reasoning. This symbolic expression is not merely interpretable: it serves as the foundation for a broad suite of patient-specific tools, including PDPs that expose local covariate effects, PRPs that summarize personalized risk contributions, feature importance metrics computed analytically from the logit matrix, and retrieval of similar cases to support case-based reasoning. Together, these tools enable both global understanding and granular, individualized insight, an essential feature for trustworthy AI in high-stakes settings like medicine. Perhaps most importantly, the availability of an analytical expression opens the door to clinical integration in ways that are rarely possible with ``black-box'' models. The \emph{KAAM’s} symbolic formula can be inspected, documented, and audited by healthcare professionals, making it possible to bridge the gap between algorithmic output and medical justification. This is not just a technical convenience: it is an ethical and legal necessity for real-world deployment.

This work introduces a versatile framework for clinical interpretability, but many exciting directions remain open for future research. 

\begin{itemize}
\item In our case-based analysis, we use Euclidean distance in logit space to retrieve similar patients. While effective, this may overlook richer geometric or probabilistic relationships. Exploring alternatives, such as Jensen–Shannon divergence or Wasserstein distance, could yield better matches, particularly in imbalanced or high-dimensional settings. 

\item Our visualization pipeline focuses on \emph{KAAMs} due to their additive structure. However, full \emph{Logistic-KANs}, which model complex feature interactions through nested function compositions, encode complex feature interactions via symbolic composition. Designing new tools to navigate these expressions (e.g., composition trees, derivative-based saliency maps, or modularity plots) could unlock their interpretability potential. 

\item While \emph{KAAMs} use B-splines to parameterize univariate functions, other basis expansions (e.g., Chebyshev polynomials, Random Fourier Features, rational splines) may offer improved approximation or interpretability depending on the domain. Studying their trade-offs is a natural next step.

\item The availability of the learned formula is a major contribution of this work, and a key differentiator from post-hoc methods. This opens avenues such as sensitivity analysis via symbolic derivatives, feature pruning via variance contribution, rule extraction or simplification, and counterfactual explanations (\textit{What was the prediction if my patients were younger?})

\item We have demonstrated an interactive prototype for clinician use. Future work should focus on integrating \emph{KAAMs} into real-time decision-support systems, exploring temporal dynamics, incorporating confidence intervals or uncertainty quantification, and tailoring interfaces to actual clinician workflows.

\item While our work focuses on structured tabular data, the \emph{KAAM} architecture is general. It could be extended to support, for example, time-series classification, multimodal fusion, or survival prediction, where transparency is critical. For instance, applying \emph{KAAMs} to longitudinal patient trajectories could yield interpretable time-varying risk estimates.

\end{itemize}

In summary, this study presents a compelling case for utilizing KAN-based models in clinical classification, particularly in scenarios that demand transparency, trust, and interpretability. By combining high performance with an interpretable symbolic core, the \emph{KAAM} offers a blueprint for AI systems that are not only powerful but also clinically usable, verifiable, and ultimately accountable.

%% file: 7_statements.tex
\section*{Author Contributions}
J.P. contributed to Conceptualization, offering leadership in defining the research goals and guiding the scientific direction of the study. S.Z. led Funding Acquisition and provided overarching Supervision throughout the project. F.F.S., A.G.L., A.A., and P.A. carried out the core Investigation and Formal Analysis, performing the experiments, analyzing data, and interpreting the results. A.A. and P.A. were responsible for Software development and for implementing the models and computational framework used in the study. A.A. also led the Methodology, designing the modeling approach and experimental pipeline. All authors contributed to Validation, ensuring the robustness and reproducibility of the findings. Additionally, all authors participated in Writing – Review \& Editing, while A.A. contributed substantially to Writing – Original Draft Preparation and Visualization of the results.

\section*{Competing Interests}
All authors declare no financial or non-financial competing interests.

\section*{Funding}
Two authors (FFS. and PAA.) receive funding from the Repo4EU project (Grant Agreement No. 101057619 \url{https://repo4.eu/}). The funder played no role in study design, data collection, analysis and interpretation of data, or the writing of this manuscript. No project-specific funding was used for this research.

\section*{Data Availability}
All datasets used in this study are publicly available. The benchmark datasets used for model development and evaluation are available in their respective public repositories, as detailed in the Methods section. In addition, all processed data and experimental configurations used to generate the results in this manuscript are openly available in \codeurl.

\section*{Reproducibility Statement}
We have made reproducibility a core focus of this work. Both proposed models, the full interpretability pipeline, and the training configurations and hyperparameter settings are thoroughly detailed in \cref{sec:methods}, \cref{sec:results}, and the appendices. All experiments rely on publicly available benchmark datasets. To support reproducibility and foster future research, we provide open access to the complete codebase, along with scripts to replicate every figure and table in the paper. These resources are intended to ensure that both the methodological and empirical aspects of our study can be independently validated and built upon.

\section*{Ethics Statement}
This study poses no immediate ethical concerns. All datasets employed are publicly available and have been preprocessed in accordance with established procedures. The primary goal of our framework is to improve model transparency and interpretability (an essential aspect for deploying AI systems responsibly in sensitive applications \citep{Rudin2018StopEB, AmannBMC2020}). By generating analytical expressions and offering direct control over the balance between performance and interpretability, the method facilitates more effective auditing, external verification, and confidence in the results. While biases in the original datasets could influence model behavior, the interpretable nature of our approach helps expose and address such issues.

\section*{LLM Usage}
Large language models (LLMs) were used exclusively for language refinement (grammar and clarity) and for assisting with the generation of plots and illustrations in Python and \LaTeX{}. All scientific ideas, methodological contributions, experimental design, implementation, and analysis were fully conceived and executed by the authors without the aid of LLMs.

%% file: 8_acknowledgements.tex
\section*{Acknowledgements}
This work was supported by the European Project: Repo4EU
\url{https://repo4.eu/} Grant no. 101057619 within the Horizon Europe Research and Innovation Programme.

%% file: 1_hyperparameter_optimization.tex
\section{Hyperparameter Optimization and Training Efficiency}
\label{sec:app:hyperparam_opt}

We performed an exhaustive hyperparameter optimization via grid search for all models. The search spaces were defined as follows:

\begin{itemize}
    \item RF: number of estimators \{20, 50\}; maximum tree depth \{10, 20\}; class balancing \{True, False\}; minimum samples per leaf \{1, 5\}.
    
    \item LR: penalty type \{L1, L2\}; class balancing \{True, False\}.
    
    \item MLP: hidden layer sizes \{32, 64, 128, 256\}; L2 regularization \{1e-3, 1e-4\}.
    
    \item NAM: number of learners \{10, 20\}; basis functions \{32, 64, 128\}; hidden layer sizes \{(64, 32), (128, 64)\}.
    
    \item \emph{Logistic-KAN}: hidden sizes \{0, 5, (5, 5)\}; B-spline grid points \{1, 3, 5\}; spline degree \{1, 3, 5\}; L1 regularization \{1e-1, 1e-2, 1e-3\}; class balancing \{True, False\}; KAN initialization \{sparse, dense\}.
    
    \item \emph{KAAM}: same as \emph{Logistic-KAN}, but limited to no hidden layers (i.e., hidden size = 0).
\end{itemize}

All configurations were evaluated using 3-fold cross-validation. The best-performing hyperparameters were selected based on their performance in validation.

Models implemented in \texttt{scikit-learn}, such as MLP, LR, and RF, are highly optimized and exhibit fast training times, with average execution times ranging from 0.06 to 0.19 seconds for MLP, 0.01 to 0.06 seconds for LR, and 0.10 to 0.16 seconds for RF. In contrast, models implemented in \texttt{PyTorch} require substantially longer training times: the NAM ranges between 76.69 and 110.62 seconds per model, the \emph{Logistic-KAN} between 8.14 and 43.64 seconds, and the \emph{KAAM} between 5.54 and 68.97 seconds. While KAN-based models are computationally more expensive than classical baselines, they remain significantly faster to train than NAMs.

%% file: 2_statistical_tests.tex
\section{Statistical Significance Analysis}
\label{sec:app:stats}

In Section \ref{sec:ml_res}, we reported the predictive performance of all models across six clinical datasets and five evaluation metrics, summarizing their overall behavior using the MRR (Table \ref{tab:res_mrr}). The MRR offers a principled, ranking-based aggregation of performance across multiple tasks and metrics, highlighting consistently strong models rather than excelling only on isolated datasets. 

To complement this ranking perspective with a formal significance test, we also conducted a Friedman test \citep{demvsar2006statistical, rainio2024evaluation} followed by Holm–Bonferroni corrected \citep{holm1979simple} pairwise Wilcoxon signed-rank tests. These non-parametric tests assess whether the observed differences between models are statistically significant across all tasks, assuming paired measurements are available. Because the Friedman test requires that all models be evaluated on the same set of tasks, NAM was excluded from this analysis due to its binary-class restriction.

Table \ref{tab:friedman_pvals} reports the adjusted p-values from these post-hoc comparisons. For each metric, the baseline model (\underline{underlined}) is the one with the highest average performance. An asterisk (*) indicates that the difference with respect to the baseline is \textit{not statistically significant} at $\alpha=0.05$.

Taken together, the MRR results and the Friedman-based p-values provide complementary evidence about model performance. The MRR highlights which models rank highest and most consistently across metrics and datasets. The Friedman test, in turn, quantifies whether those differences are statistically meaningful rather than due to chance. For instance, \emph{Logistic-KAN} achieves the highest overall MRR (0.71), reflecting consistent top-ranked performance across tasks. The Friedman post-hoc tests show that the difference between \emph{Logistic-KAN} and \emph{KAAM} is generally not statistically significant, indicating comparable performance between these two interpretable models. Conversely, neural baselines such as MLP or NAM exhibit lower MRR values and significantly different (worse) performance according to the Friedman test, reinforcing the robustness of our proposed approaches.

This two-tier evaluation  (ranking plus significance testing) offers a more comprehensive and reliable picture of model performance than either measure alone, underscoring the practical advantage of \emph{Logistic-KAN} and \emph{KAAM} in tabular clinical prediction tasks.

\renewcommand{\arraystretch}{1.35}
\begin{table}[!ht]
    \centering
    \resizebox{0.5\linewidth}{!}{%
    \begin{tabular}{l|ccccc}
    \hline\hline
\textbf{Classification}     & \multicolumn{5}{c}{\textbf{Model}}     \\
    
     \textbf{Metric}  &   \textbf{MLP} &   \textbf{LR} &   \textbf{RF} &   \textbf{Logistic-KAN} &   \textbf{KAAM} \\
       
    \hline\hline
     \textit{Accuracy}      &  0.170* & 0.288* & 0.876* & \underline{1.000*} & 1.000*\\
     \textit{ROC-AUC}       &  0.006  & 0.876* & 0.631* & 0.876* & \underline{0.876*} \\
     \textit{F1-score}            &  0.026  & 1.000* & \underline{1.000*} & 1.000* & 1.000* \\
     \textit{Precision}     &  0.044  & 1.000* & \underline{1.000*} & 1.000* & 1.000* \\
     \textit{Recall}        &  0.089* & 0.256* & 0.241* & \underline{1.000*} & 1.000* \\
    \hline
    \textbf{\textit{All metrics}}    & \ensuremath{<}1e-3 & 0.068* & 0.109* & \underline{1.000*} & 1.000* \\
    \hline\hline
    \end{tabular}
    }

    \caption{Adjusted p-values from the Holm–Bonferroni corrected post-hoc comparisons of each model against the best-performing \underline{baseline} for each metric. Values correspond to the corrected p-value of the paired Wilcoxon test. An asterisk (*) indicates that the difference with respect to the baseline is \textit{not statistically significant} at $\alpha=0.05$, i.e., the model performs comparably to the baseline for that metric. Lower p-values (without *) indicate statistically significant differences (worse performance).}
    \label{tab:friedman_pvals}
\end{table}

%% file: 3_more_results.tex
\section{Additional Results}
\label{sec:app:more_results}

This appendix presents the complete symbolic expression learned by the \emph{KAAM} model for the Diabetes-130 dataset, a multiclass diagnostic classification task. Each equation corresponds to the logit $l_k$ associated with class $k$, following the additive structure defined in Equation \eqref{eq:asf_kan}.

{\footnotesize
\begin{equation}
    \label{eq:formula_4_0}
    \begin{split}
    l_0 =& - 0.05 \times \mathrm{age} - 0.143 \times \mathrm{change} + 0.035 \times \mathrm{diabetesMed} \\
    \quad&- 0.023 \times \mathrm{diag_{3}} + 0.015 \times \mathrm{gender} \\
    \quad&+ 0.158 \times \mathrm{glyburide-metformin} - 0.263 \times \mathrm{nateglinide} \\
    \quad&- 0.01 \times \mathrm{num_{lab procedures}} - 0.055 \times \mathrm{number_{emergency}} \\ 
    \quad&- 0.056 \times \mathrm{number_{outpatient}} + 0.032 \times \left(0.979 - \mathrm{repaglinide}\right)^{2} \\
    \quad&+ 0.088 \times \left(1 - 0.563 \times \mathrm{rosiglitazone}\right)^{2} \\
    \quad& - 0.243 \times \sin{\left(0.373 \times \mathrm{admission_{type id}} + 5.387 \right)} \\
    \quad& - 0.407 \times \sin{\left(5.64 \times \mathrm{glimepiride} + 1.377 \right)} \\
    \quad& + 0.487 \times \sin{\left(5.754 \times \mathrm{glipizide} - 1.966 \right)} \\
    \quad& - 0.261 \times \sin{\left(0.383 \times \mathrm{num_{medications}} + 5.188 \right)} \\
    \quad& - 0.293 \times \sin{\left(0.544 \times \mathrm{pioglitazone} + 8.238 \right)} \\
    \quad& - 0.129 \times \sin{\left(0.329 \times \mathrm{time_{in hospital}} - 7.433 \right)} \\
    \quad& + 0.097 \times \cos{\left(0.752 \times \mathrm{admission_{source id}} + 0.583 \right)} \\
    \quad& + 0.493 \times \cos{\left(5.816 \times \mathrm{insulin} + 5.997 \right)} \\
    \quad& + 0.048 \times \cos{\left(5.634 \times \mathrm{metformin} - 9.613 \right)} \\
    \quad&- 0.04 \tan{\left(0.633 \times \mathrm{glyburide} - 1.217 \right)} \\
    \quad& + 0.012 \times \left|{9.318 \times \mathrm{race} - 2.978}\right| + 0.947 \\
    \quad& - 0.041 \times e^{- 4.184 \times \mathrm{tolbutamide}} \\
    \quad& - \frac{0.506}{\sqrt{0.328 \mathrm{num_{procedures}} + 1}} 
\end{split}
\end{equation}

\begin{equation}
    \label{eq:formula_4_1}
    \begin{split}
    l_1 =& - 0.112 \times \mathrm{age} + 0.045 \times \mathrm{change} + 0.091 \times \mathrm{diabetesMed} \\
    \quad& - 0.076 \times \mathrm{discharge_{disposition id}} + 0.063 \times \mathrm{gender} \\
    \quad& - 0.236 \times \mathrm{glyburide-metformin} - 0.039 \times \mathrm{nateglinide} \\
    \quad& + 0.189 \times \mathrm{number_{inpatient}} + 0.064 \times \mathrm{number_{outpatient}} \\
    \quad& - 0.112 \times \mathrm{tolbutamide} + 0.007 \times \sqrt{\mathrm{glipizide} + 0.374} \\
    \quad& + 0.148 \times \left(- 0.46 \times \mathrm{repaglinide} - 1\right)^{2} \\
    \quad& + 0.15 \times \left(- 0.348 \times \mathrm{rosiglitazone} - 1\right)^{2} \\
    \quad& + 0.012 \times \log{\left(9.993 \times \mathrm{metformin} + 0.156 \right)} \\
    \quad& + 0.079 \times \log{\left(8.436 \times \mathrm{race} + 2.158 \right)} \\
    \quad&- 0.686 \times \sin{\left(0.409 \times \mathrm{glimepiride} + 2.178 \right)} \\
    \quad& - 0.064 \times \sin{\left(1.024 \times \mathrm{glyburide} + 4.192 \right)} \\
    \quad& + 0.027 \times \tan{\left(2.587 \times \mathrm{insulin} + 0.44 \right)} \\
    \quad& + 0.145 + \frac{0.005}{0.51 \times - 1.207 \times \mathrm{pioglitazone}} 
\end{split}
\end{equation}

\begin{equation}
    \label{eq:formula_4_2}
    \begin{split}
    l_2 =& 0.018 \times \mathrm{admission_{type id}} - 0.092 \times \mathrm{age} + 0.018 \times \mathrm{change} \\
    \quad& + 0.009 \times \mathrm{diabetesMed} - 0.076 \times \mathrm{gender} \\
    \quad& + 0.295 \times \mathrm{nateglinide} - 0.089 \times \mathrm{num_{procedures}} \\
    \quad&+ 0.107 \times \mathrm{number_{emergency}} + 0.112 \times \mathrm{number_{inpatient}} \\
    \quad&+ 0.064 \times \mathrm{number_{outpatient}} \\
    \quad& + 0.059 \times \left(- 0.183 \times \mathrm{glimepiride} - 1\right)^{2}  \\
    \quad& + 0.319 \times \left(- 0.061 \times \mathrm{glyburide} - 1\right)^{2} \\
    \quad&  - 0.024 \times \left(- \mathrm{repaglinide} - 0.775\right)^{2} \\
    \quad& + 0.184 \times \left(- 0.294 \times \mathrm{rosiglitazone} - 1\right)^{2} \\
    \quad&  - 0.462 \times e^{0.06 \times \mathrm{metformin}}  \\
    \quad& + 0.338 \times \sin{\left(0.482 \times \mathrm{admission_{source id}} - 1.261 \right)} \\
    \quad& - 0.359 \times \sin{\left(6.775 \times \mathrm{glipizide} - 7.388 \right)} \\
    \quad& + 0.151 \times \sin{\left(0.366 \times \mathrm{race} - 7.365 \right)}\\
    \quad& - 0.097 \times \cos{\left(0.38 \times \mathrm{diag_{3}} + 0.774 \times \right)} \\
    \quad& - 0.335 \times \cos{\left(0.402 \times \mathrm{discharge_{disposition id}} - 5.804 \right)}\\
    \quad& - 0.146 \times \cos{\left(0.96 \times \mathrm{insulin} - 3.135 \right)} \\
    \quad& - 0.363 \times \cos{\left(0.346 \times \mathrm{num_{medications}} - 5.816 \right)} \\
    \quad& + 0.317 \times \cos{\left(0.344 \times \mathrm{time_{in hospital}} + 6.769 \right)} \\
    \quad& - 0.339 + 0.024 \times e^{- 8.408 \times \mathrm{tolbutamide}} \\
    \quad& + 0.011 \times e^{- 4.491 \times \mathrm{glyburide-metformin}} \\
    \quad& - \frac{0.002}{0.162 - 1.027 \times \mathrm{pioglitazone}}.
\end{split}
\end{equation}
}

The performance metrics obtained from this symbolic \emph{KAAM} model (using the formulas above) closely match those of the original \emph{KAAM} model trained with full B-splines and without pruning. Specifically, the symbolic version achieves accuracy of $0.57/0.58$, precision of $0.50/0.50$, recall of $0.57/0.58$, F1-score of $0.48/0.50$, and ROC-AUC of $0.56/0.54$, demonstrating that expressivity is retained while gaining interpretability.

%% file: main.bib
@article{xie2025evolution,
  title={Evolution of artificial intelligence in healthcare: a 30-year bibliometric study},
  author={Xie, Yaojue and Zhai, Yuansheng and Lu, Guihua},
  journal={Frontiers in Medicine},
  volume={11},
  pages={1505692},
  year={2025},
  publisher={Frontiers Media SA}
}

@article{bindra2024artificial,
  title={Artificial intelligence in medical science: a review},
  author={Bindra, Simrata and Jain, Richa},
  journal={Irish Journal of Medical Science (1971-)},
  volume={193},
  number={3},
  pages={1419--1429},
  year={2024},
  publisher={Springer}
}

@article{xu2024medical,
  title={Medical artificial intelligence and the black box problem: a view based on the ethical principle of “do no harm”},
  author={Xu, Hanhui and Shuttleworth, Kyle Michael James},
  journal={Intelligent Medicine},
  volume={4},
  number={1},
  pages={52--57},
  year={2024},
  publisher={Elsevier}
}

@article{ghassemi2021false,
  title={The false hope of current approaches to explainable artificial intelligence in health care},
  author={Ghassemi, Marzyeh and Oakden-Rayner, Luke and Beam, Andrew L},
  journal={The Lancet Digital Health},
  volume={3},
  number={11},
  pages={e745--e750},
  year={2021},
  publisher={Elsevier}
}

@article{frasca2024explainable,
  title={Explainable and interpretable artificial intelligence in medicine: a systematic bibliometric review},
  author={Frasca, Maria and La Torre, Davide and Pravettoni, Gabriella and Cutica, Ilaria},
  journal={Discover Artificial Intelligence},
  volume={4},
  number={1},
  pages={15},
  year={2024},
  publisher={Springer}
}

@article{lundberg2017unified,
  title={A unified approach to interpreting model predictions},
  author={Lundberg, Scott M and Lee, Su-In},
  journal={Advances in neural information processing systems},
  volume={30},
  year={2017}
}

@misc{mosca2022shap,
  title={SHAP-based explanation methods: a review for NLP interpretability},
  author={Mosca, Edoardo and Szigeti, Ferenc and Tragianni, Stella and Gallagher, Daniel and Groh, Georg},
  booktitle={Proceedings of the 29th international conference on computational linguistics},
  pages={4593--4603},
  year={2022}
}

@misc{ribeiro2016should,
  title={"Why should I trust you?" Explaining the predictions of any classifier},
  author={Ribeiro, Marco Tulio and Singh, Sameer and Guestrin, Carlos},
  booktitle={Proceedings of the 22nd ACM SIGKDD international conference on knowledge discovery and data mining},
  pages={1135--1144},
  year={2016}
}

@article{covert2020understanding,
  title={Understanding global feature contributions with additive importance measures},
  author={Covert, Ian and Lundberg, Scott M and Lee, Su-In},
  journal={Advances in Neural Information Processing Systems},
  volume={33},
  pages={17212--17223},
  year={2020}
}

@article{selvaraju2020grad,
  title={Grad-CAM: visual explanations from deep networks via gradient-based localization},
  author={Selvaraju, Ramprasaath R and Cogswell, Michael and Das, Abhishek and Vedantam, Ramakrishna and Parikh, Devi and Batra, Dhruv},
  journal={International journal of computer vision},
  volume={128},
  pages={336--359},
  year={2020},
  publisher={Springer}
}

@article{alaa2019demystifying,
  title={Demystifying black-box models with symbolic metamodels},
  author={Alaa, Ahmed M and Van der Schaar, Mihaela},
  journal={Advances in neural information processing systems},
  volume={32},
  year={2019}
}

@article{lipton2018mythos,
  title={The mythos of model interpretability: In machine learning, the concept of interpretability is both important and slippery.},
  author={Lipton, Zachary C},
  journal={Queue},
  volume={16},
  number={3},
  pages={31--57},
  year={2018},
  publisher={ACM New York, NY, USA}
}

@article{chen2022explainable,
  title={Explainable medical imaging AI needs human-centered design: guidelines and evidence from a systematic review},
  author={Chen, Haomin and Gomez, Catalina and Huang, Chien-Ming and Unberath, Mathias},
  journal={NPJ digital medicine},
  volume={5},
  number={1},
  pages={156},
  year={2022},
  publisher={Nature Publishing Group UK London}
}

@article{pahud2024orchestrating,
  title={Orchestrating explainable artificial intelligence for multimodal and longitudinal data in medical imaging},
  author={Pahud de Mortanges, Aur{\'e}lie and Luo, Haozhe and Shu, Shelley Zixin and Kamath, Amith and Suter, Yannick and Shelan, Mohamed and P{\"o}llinger, Alexander and Reyes, Mauricio},
  journal={NPJ digital medicine},
  volume={7},
  number={1},
  pages={195},
  year={2024},
  publisher={Nature Publishing Group UK London}
}

@article{alkhanbouli2025role,
  title={The role of explainable artificial intelligence in disease prediction: a systematic literature review and future research directions},
  author={Alkhanbouli, Razan and Matar Abdulla Almadhaani, Hour and Alhosani, Farah and Simsekler, Mecit Can Emre},
  journal={BMC Medical Informatics and Decision Making},
  volume={25},
  number={1},
  pages={110},
  year={2025},
  publisher={Springer}
}

@article{bienefeld2023solving,
  title={Solving the explainable AI conundrum by bridging clinicians’ needs and developers’ goals},
  author={Bienefeld, Nadine and Boss, Jens Michael and L{\"u}thy, Rahel and Brodbeck, Dominique and Azzati, Jan and Blaser, Mirco and Willms, Jan and Keller, Emanuela},
  journal={NPJ Digital Medicine},
  volume={6},
  number={1},
  pages={94},
  year={2023},
  publisher={Nature Publishing Group UK London}
}

@article{liu2024kan,
  title={Kan: Kolmogorov-arnold networks},
  author={Liu, Ziming and Wang, Yixuan and Vaidya, Sachin and Ruehle, Fabian and Halverson, James and Solja{\v{c}}i{\'c}, Marin and Hou, Thomas Y and Tegmark, Max},
  journal={arXiv preprint arXiv:2404.19756},
  year={2024}
}

@article{liu2024kan2,
  title={Kan 2.0: Kolmogorov-arnold networks meet science},
  author={Liu, Ziming and Ma, Pingchuan and Wang, Yixuan and Matusik, Wojciech and Tegmark, Max},
  journal={arXiv preprint arXiv:2408.10205},
  year={2024}
}

@article{yu2024kan,
  title={Kan or mlp: A fairer comparison},
  author={Yu, Runpeng and Yu, Weihao and Wang, Xinchao},
  journal={arXiv preprint arXiv:2407.16674},
  year={2024}
}

@article{altarabichi2024dropkan,
  title={Dropkan: Regularizing kans by masking post-activations},
  author={Altarabichi, Mohammed Ghaith},
  journal={arXiv preprint arXiv:2407.13044},
  year={2024}
}

@article{yu2024residual,
  title={Residual kolmogorov-arnold network for enhanced deep learning},
  author={Yu, Ray Congrui and Wu, Sherry and Gui, Jiang},
  journal={arXiv preprint arXiv:2410.05500},
  year={2024}
}

@article{genet2024temporal,
  title={A temporal kolmogorov-arnold transformer for time series forecasting},
  author={Genet, Remi and Inzirillo, Hugo},
  journal={ArXiv},
  year={2024}
}

@article{bodner2024convolutional,
  title={Convolutional kolmogorov-arnold networks},
  author={Bodner, Alexander Dylan and Tepsich, Antonio Santiago and Spolski, Jack Natan and Pourteau, Santiago},
  journal={arXiv preprint arXiv:2406.13155},
  year={2024}
}

@article{drokin2024kolmogorov,
  title={Kolmogorov-arnold convolutions: Design principles and empirical studies},
  author={Drokin, Ivan},
  journal={arXiv preprint arXiv:2407.01092},
  year={2024}
}

@article{li2024u,
  title={U-kan makes strong backbone for medical image segmentation and generation},
  author={Li, Chenxin and Liu, Xinyu and Li, Wuyang and Wang, Cheng and Liu, Hengyu and Liu, Yifan and Chen, Zhen and Yuan, Yixuan},
  journal={arXiv preprint arXiv:2406.02918},
  year={2024}
}

@article{zhang2024graphkan,
  title={Graphkan: Enhancing feature extraction with graph kolmogorov arnold networks},
  author={Zhang, Fan and Zhang, Xin},
  journal={arXiv preprint arXiv:2406.13597},
  year={2024}
}

@article{chen2024gaussian,
  title={Gaussian Process Kolmogorov-Arnold Networks},
  author={Chen, Andrew Siyuan},
  journal={arXiv preprint arXiv:2407.18397},
  year={2024}
}

@misc{kich2024kolmogorov,
  title={Kolmogorov-Arnold Networks for Online Reinforcement Learning},
  author={Kich, Victor A and Bottega, Jair A and Steinmetz, Raul and Grando, Ricardo B and Yorozu, Ayano and Ohya, Akihisa},
  booktitle={2024 24th International Conference on Control, Automation and Systems (ICCAS)},
  pages={958--963},
  year={2024},
  organization={IEEE}
}

@article{ji2024comprehensive,
  title={A comprehensive survey on kolmogorov arnold networks (kan)},
  author={Ji, Tianrui and Hou, Yuntian and Zhang, Di},
  journal={arXiv preprint arXiv:2407.11075},
  year={2024}
}

@article{somvanshi2024survey,
  title={A survey on kolmogorov-arnold network},
  author={Somvanshi, Shriyank and Javed, Syed Aaqib and Islam, Md Monzurul and Pandit, Diwas and Das, Subasish},
  journal={arXiv preprint arXiv:2411.06078},
  year={2024}
}

@article{yang2025medkan,
  title={Medkan: An advanced kolmogorov-arnold network for medical image classification},
  author={Yang, Zhuoqin and Zhang, Jiansong and Luo, Xiaoling and Lu, Zheng and Shen, Linlin},
  journal={arXiv preprint arXiv:2502.18416},
  year={2025}
}

@article{agarwal2021neural,
  title={Neural additive models: Interpretable machine learning with neural nets},
  author={Agarwal, Rishabh and Melnick, Levi and Frosst, Nicholas and Zhang, Xuezhou and Lengerich, Ben and Caruana, Rich and Hinton, Geoffrey E},
  journal={Advances in neural information processing systems},
  volume={34},
  pages={4699--4711},
  year={2021}
}

@article{hornik1989multilayer,
  title={Multilayer feedforward networks are universal approximators},
  author={Hornik, Kurt and Stinchcombe, Maxwell and White, Halbert},
  journal={Neural networks},
  volume={2},
  number={5},
  pages={359--366},
  year={1989},
  publisher={Elsevier}
}

@book{haykin1994neural,
  title={Neural networks: a comprehensive foundation},
  author={Haykin, Simon},
  year={1994},
  publisher={Prentice Hall PTR}
}

@book{kolmogorov1961representation,
  title={On the representation of continuous functions of several variables by superpositions of continuous functions of a smaller number of variables},
  author={Kolmogorov, Andre{\u\i} Nikolaevich},
  year={1961},
  publisher={American Mathematical Society}
}

@article{poggio2020theoretical,
  title={Theoretical issues in deep networks},
  author={Poggio, Tomaso and Banburski, Andrzej and Liao, Qianli},
  journal={Proceedings of the National Academy of Sciences},
  volume={117},
  number={48},
  pages={30039--30045},
  year={2020},
  publisher={National Academy of Sciences}
}

@article{girosi1989representation,
  title={Representation properties of networks: Kolmogorov's theorem is irrelevant},
  author={Girosi, Federico and Poggio, Tomaso},
  journal={Neural Computation},
  volume={1},
  number={4},
  pages={465--469},
  year={1989},
  publisher={MIT Press}
}

@article{kuurkova1991kolmogorov,
  title={Kolmogorov's theorem is relevant},
  author={Kurkova, Vera},
  journal={Neural computation},
  volume={3},
  number={4},
  pages={617--622},
  year={1991},
  publisher={MIT Press}
}

@article{schmidt2021kolmogorov,
  title={The Kolmogorov--Arnold representation theorem revisited},
  author={Schmidt-Hieber, Johannes},
  journal={Neural networks},
  volume={137},
  pages={119--126},
  year={2021},
  publisher={Elsevier}
}

@book{hastie1990generalized,
  title={Generalized Additive Models},
  author={Hastie, TJ and Tibshirani, RJ},
  volume={43},
  year={1990},
  publisher={CRC Press}
}

@article{marcinkevivcs2023interpretable,
  title={Interpretable and explainable machine learning: A methods-centric overview with concrete examples},
  author={Marcinkevi{\v{c}}s, Ri{\v{c}}ards and Vogt, Julia E},
  journal={Wiley Interdisciplinary Reviews: Data Mining and Knowledge Discovery},
  volume={13},
  number={3},
  pages={e1493},
  year={2023},
  publisher={Wiley Online Library}
}

@article{segal1994sufficient,
  title={A sufficient condition for additively separable functions},
  author={Segal, Uzi},
  journal={Journal of Mathematical Economics},
  volume={23},
  number={3},
  pages={295--303},
  year={1994},
  publisher={Elsevier}
}

@misc{lou2012intelligible,
  title={Intelligible models for classification and regression},
  author={Lou, Yin and Caruana, Rich and Gehrke, Johannes},
  booktitle={Proceedings of the 18th ACM SIGKDD international conference on Knowledge discovery and data mining},
  pages={150--158},
  year={2012}
}

@article{hastie1995generalized,
  title={Generalized additive models for medical research.},
  author={Hastie, T and Tibshirani, R},
  journal={Statistical Methods in Medical Research},
  volume={4},
  number={3},
  pages={187--196},
  year={1995}
}

@article{utkin2022survnam,
  title={SurvNAM: The machine learning survival model explanation},
  author={Utkin, Lev V and Satyukov, Egor D and Konstantinov, Andrei V},
  journal={Neural Networks},
  volume={147},
  pages={81--102},
  year={2022},
  publisher={Elsevier}
}

@article{fasiolo2020scalable,
  title={Scalable visualization methods for modern generalized additive models},
  author={Fasiolo, Matteo and Nedellec, Rapha{\"e}l and Goude, Yannig and Wood, Simon N},
  journal={Journal of computational and Graphical Statistics},
  volume={29},
  number={1},
  pages={78--86},
  year={2020},
  publisher={Taylor \& Francis}
}

@misc{hohman2019telegam,
  title={TeleGam: Combining visualization and verbalization for interpretable machine learning},
  author={Hohman, Fred and Srinivasan, Arjun and Drucker, Steven M},
  booktitle={2019 IEEE Visualization Conference (VIS)},
  pages={151--155},
  year={2019},
  organization={IEEE}
}

@article{wang2021gam,
  title={Gam changer: Editing generalized additive models with interactive visualization},
  author={Wang, Zijie J and Kale, Alex and Nori, Harsha and Stella, Peter and Nunnally, Mark and Chau, Duen Horng and Vorvoreanu, Mihaela and Vaughan, Jennifer Wortman and Caruana, Rich},
  journal={arXiv preprint arXiv:2112.03245},
  year={2021}
}

@misc{caruana2015intelligible,
  title={Intelligible models for healthcare: Predicting pneumonia risk and hospital 30-day readmission},
  author={Caruana, Rich and Lou, Yin and Gehrke, Johannes and Koch, Paul and Sturm, Marc and Elhadad, Noemie},
  booktitle={Proceedings of the 21th ACM SIGKDD international conference on knowledge discovery and data mining},
  pages={1721--1730},
  year={2015}
}

@book{hosmer2013applied,
  title={Applied logistic regression},
  author={Hosmer Jr, David W and Lemeshow, Stanley and Sturdivant, Rodney X},
  year={2013},
  publisher={John Wiley \& Sons}
}

@article{cox1972regression,
  title={Regression models and life-tables},
  author={Cox, David R},
  journal={Journal of the Royal Statistical Society: Series B (Methodological)},
  volume={34},
  number={2},
  pages={187--202},
  year={1972},
  publisher={Wiley Online Library}
}

@article{garrido2014methods,
  title={Methods for constructing and assessing propensity scores},
  author={Garrido, Melissa M and Kelley, Amy S and Paris, Julia and Roza, Katherine and Meier, Diane E and Morrison, R Sean and Aldridge, Melissa D},
  journal={Health services research},
  volume={49},
  number={5},
  pages={1701--1720},
  year={2014},
  publisher={Wiley Online Library}
}

@article{rosenbaum83propensity,
 ISSN = {00063444, 14643510},
  author = {Paul R. Rosenbaum and Donald B. Rubin},
 journal = {Biometrika},
 number = {1},
 pages = {41--55},
 publisher = {[Oxford University Press, Biometrika Trust]},
 title = {The Central Role of the Propensity Score in Observational Studies for Causal Effects},
 urldate = {2025-07-29},
 volume = {70},
 year = {1983}
}

@article{austin2011introduction,
  title={An introduction to propensity score methods for reducing the effects of confounding in observational studies},
  author={Austin, Peter C},
  journal={Multivariate behavioral research},
  volume={46},
  number={3},
  pages={399--424},
  year={2011},
  publisher={Taylor \& Francis}
}

@article{austin2010performance,
  title={The performance of different propensity-score methods for estimating differences in proportions (risk differences or absolute risk reductions) in observational studies},
  author={Austin, Peter C},
  journal={Statistics in medicine},
  volume={29},
  number={20},
  pages={2137--2148},
  year={2010},
  publisher={Wiley Online Library}
}

@book{hernan2020causal,
  title     = {Causal Inference: What If},
  author    = {Hern{\'a}n, Miguel A. and Robins, James M.},
  year      = {2020},
  publisher = {Chapman \& Hall/CRC},
  address   = {Boca Raton, FL},
}

@misc{hastie2009elements,
  title={The elements of statistical learning},
  author={Hastie, Trevor and Tibshirani, Robert and Friedman, Jerome and others}
}

@article{saary2008radar,
  title={Radar plots: a useful way for presenting multivariate health care data},
  author={Saary, M Joan},
  journal={Journal of clinical epidemiology},
  volume={61},
  number={4},
  pages={311--317},
  year={2008},
  publisher={Elsevier}
}

@article{xie2019building,
  title={Building risk prediction models for type 2 diabetes using machine learning techniques},
  author={Xie, Zidian and Nikolayeva, Olga and Luo, Jiebo and Li, Dongmei},
  journal={Preventing chronic disease},
  volume={16},
  pages={E130},
  year={2019}
}

@misc{diabetes_130-us_hospitals_for_years_1999-2008_296,
  author       = {Clore, John and Cios, Krzysztof and DeShazo, Jon and Strack, Beata},
  title        = {{Diabetes 130-US Hospitals for Years 1999-2008}},
  year         = {2014},
  howpublished = {UCI Machine Learning Repository},
}

@misc{cdc_brfss_2022,
  author       = {{Centers for Disease Control and Prevention (CDC)}},
  title        = {{Behavioral Risk Factor Surveillance System (BRFSS) 2022 Data}},
  year         = {2023},
  note         = {Accessed: August 13, 2025}
}

@article{palechor2019dataset,
  title={Dataset for estimation of obesity levels based on eating habits and physical condition in individuals from Colombia, Peru and Mexico},
  author={Palechor, Fabio Mendoza and De la Hoz Manotas, Alexis},
  journal={Data in brief},
  volume={25},
  pages={104344},
  year={2019},
  publisher={Elsevier}
}

@article{weinstein2013cancer,
  title={The cancer genome atlas pan-cancer analysis project},
  author={Weinstein, John N and Collisson, Eric A and Mills, Gordon B and Shaw, Kenna R and Ozenberger, Brad A and Ellrott, Kyle and Shmulevich, Ilya and Sander, Chris and Stuart, Joshua M},
  journal={Nature genetics},
  volume={45},
  number={10},
  pages={1113--1120},
  year={2013},
  publisher={Nature Publishing Group}
}

@article{loftus2024causal,
  title={Causal dependence plots},
  author={Loftus, Joshua and Bynum, Lucius and Hansen, Sakina},
  journal={Advances in Neural Information Processing Systems},
  volume={37},
  pages={112656--112683},
  year={2024}
}

@article{demvsar2006statistical,
  title={Statistical comparisons of classifiers over multiple data sets},
  author={Dem{\v{s}}ar, Janez},
  journal={Journal of Machine learning research},
  volume={7},
  number={Jan},
  pages={1--30},
  year={2006}
}

@article{rainio2024evaluation,
  title={Evaluation metrics and statistical tests for machine learning},
  author={Rainio, Oona and Teuho, Jarmo and Kl{\'e}n, Riku},
  journal={Scientific Reports},
  volume={14},
  number={1},
  pages={6086},
  year={2024},
  publisher={Nature Publishing Group UK London}
}

@article{holm1979simple,
  title={A simple sequentially rejective multiple test procedure},
  author={Holm, Sture},
  journal={Scandinavian journal of statistics},
  pages={65--70},
  year={1979},
  publisher={JSTOR}
}

@article{Rudin2018StopEB,
  title={Stop explaining black box machine learning models for high stakes decisions and use interpretable models instead},
  author={Cynthia Rudin},
  journal={Nature Machine Intelligence},
  year={2018},
  volume={1},
  pages={206 - 215},
}

@article{AmannBMC2020,
  author  = {Amann, Julia and Blasimme, Alessandro and Vayena, Effy and Frey, Dietmar and Madai, Vince I.},
  title   = {Explainability for artificial intelligence in healthcare: a multidisciplinary perspective},
  journal = {BMC Medical Informatics and Decision Making},
  year    = {2020},
  volume  = {20},
  number  = {310},
  doi     = {10.1186/s12911-020-01332-6}
}

@article{wachter2017counterfactual,
  title={Counterfactual explanations without opening the black box: Automated decisions and the GDPR},
  author={Wachter, Sandra and Mittelstadt, Brent and Russell, Chris},
  journal={Harv. JL \& Tech.},
  volume={31},
  pages={841},
  year={2017},
  publisher={HeinOnline}
}

@book{pearl2009causality,
  title={Causality},
  author={Pearl, Judea},
  year={2009},
  publisher={Cambridge university press}
}
